# AbsorbEvo: An Agentic Framework for Autonomous Inverse Design of Microwave Absorbers

Zhicheng Feng[1], Yubo Zhao[1], Xuefeng Yao[1]*

[1]*Department of Engineering Mechanics, Applied Mechanics Lab, Tsinghua University, Beijing 100084, P. R. China*

*Corresponding author. Email: yxf@mail.tsinghua.edu.cn

**Abstract**

Designing high-performance microwave absorbers requires specialized expertise in electromagnetic theory, materials science and simulation programming, and entails time-consuming optimization. Here, we present AbsorbEvo, an agentic framework for autonomous inverse design that translates natural-language performance objectives into designs verified by full-wave simulations. Its candidate evolution strategy integrates language reasoning, physics-based prediction and historical feedback. A large language model proposes the directions and magnitudes of parameter adjustments based on task objectives and computational history. The system combines directed increments with global sampling to generate candidates and uses a low-cost predictive model as a physics prior to rank them. Only high-ranking designs undergo full-wave simulation. Results passing physical validity checks are used to evaluate performance and guide subsequent search. Experience from training tasks is further distilled into textual skills, which are independently validated before use in new tasks. Under identical proposal budgets on held-out AbsorbBench-36 tasks, AbsorbEvo achieved a task success rate of 79.17%, versus 25.00% for a generic agent and 12.50% for random search. Its mean best coverage was 0.7816, compared with 0.6434 and 0.6448, respectively. By integrating language reasoning and physics-based feedback into design decisions, AbsorbEvo provides a methodological foundation for natural-language-driven autonomous inverse design of microwave absorbers.

**Keywords:** microwave absorbers; inverse design; large language model agents; multi-fidelity optimization; physics verification

## Introduction

Microwave absorbers suppress electromagnetic reflection through material losses and geometric tailoring, with important applications in radar stealth, electromagnetic compatibility and interference mitigation in wireless systems [1]. In complex electromagnetic environments, absorbers must achieve efficient absorption within target frequency bands and, in many applications, also satisfy thickness constraints and performance requirements across different incidence angles and polarizations [2, 3]. Their response depends on the interplay of material dispersion, impedance matching and structural resonances [4]. Increasing material losses or adjusting local dimensions alone is therefore often insufficient to meet the full set of design requirements. Jointly determining material and geometric parameters from a target electromagnetic response is thus an important route to high-performance absorbers.

Absorber geometries can be represented explicitly or implicitly. Explicit representations define geometry directly through dimensions, boundaries and relationships between components, whereas implicit representations use spatial functions and their isosurfaces to define structural form and the distribution of solid material. Despite these different modelling approaches, both require linking design variables to electromagnetic responses and searching for feasible designs while accounting for coupling among material properties, geometry and incidence conditions. Autonomous design therefore extends beyond parameter optimization: it requires translating design requirements into executable models and iteratively updating designs using reliable computational feedback.

Coupling optimization algorithms with electromagnetic simulation platforms enables simultaneous optimization of multiple structural parameters through high-fidelity full-wave evaluation, but repeated simulations often incur high computational costs and long optimization cycles. Theoretical models, including those based on effective material parameters, can rapidly estimate reflection responses at substantially lower cost. However, approximation errors make discrepancies between the predicted performance of optimized designs and full-wave or experimental results difficult to avoid. Data-driven surrogate models accelerate performance prediction but typically require substantial training data and lack reliable extrapolation to parameter regions not covered by those data [5, 6]. These methods offer distinct advantages in computational efficiency, predictive accuracy and applicability, yet model selection, tool configuration and search strategy adjustment still depend on expert knowledge. Iterative optimization thus remains cumbersome and time-consuming, particularly in large joint material–geometry parameter spaces. Integrating these evaluation methods and automating the design workflow therefore offer an important route to efficient optimization with reliable results.

Artificial intelligence is expanding beyond data analysis and performance prediction to generating solutions and organizing research workflows [7]. Large language model (LLM) agents offer a new way to connect natural-language requirements, domain knowledge and scientific computing tools [8, 9]. In materials discovery, Ghafarollahi and Buehler introduced SciAgents, which combines knowledge graphs with multi-agent reasoning to generate and iteratively refine research hypotheses for bioinspired materials [10]. In computational mechanics, Ni and Buehler proposed MechAgents, which uses multi-agent collaboration for planning, finite-element code generation, execution and error correction in elasticity problems, linking language reasoning with physics-based modelling [11]. Deotale et al. introduced ALL-FEM, which combines a language model fine-tuned for FEniCS with a multi-agent workflow for problem formulation, finite-element code generation, execution, debugging and result visualization [12]. These studies show that agents have evolved beyond coding assistance to organizing and executing complete engineering analysis workflows.

In photonic device design, agents also participate in model construction and inverse optimization. Lu et al. proposed an agentic framework for photonic metamaterials that autonomously builds forward predictive models, invokes optimization tools and performs inverse design based on target spectra, translating design intent into structural designs [13]. Lupoiu et al. introduced MetaChat, which combines multi-agent reasoning with a Maxwell surrogate solver for multi-objective, multi-wavelength design of freeform metasurfaces [14]. These studies link requirement interpretation, model invocation and design search, highlighting the potential for agents to autonomously adapt the design process using intermediate feedback. For cross-task experience reuse, Huang et al. combined human-readable external skill files with a fixed physics solver and an independent evaluator. Computational feedback guides skill revision, while validation tasks determine which versions are adopted for subsequent use, enabling adaptation in metasurface inverse design through textual guidance rather than changes to model weights [15].

Reliable modelling and simulation underpin absorber design across frequency bands, incidence angles and polarizations. A greater challenge is jointly determining material and geometric parameters from a target electromagnetic response. Language models can organize design steps, but their outputs may contain errors and require verification against independent evidence [16, 17]. Autonomous inverse design must combine physics-based predictions of potential improvements with full-wave feedback from previous designs to identify effective adjustments and refine search directions, allowing subsequent exploration to build on prior attempts.

To this end, we introduce AbsorbEvo, an AI-driven inverse design agent that translates natural-language performance objectives into absorber designs. At its core is a candidate evolution strategy jointly driven by language reasoning and physics-based prediction. A large language model proposes the directions and magnitudes of parameter adjustments based on task requirements and search history. The system generates candidate designs along these directions while incorporating global sampling to maintain search diversity. A physics prior based on effective-medium theory and wave propagation in layered media compares the expected performance of candidates, and designs with high predicted performance are submitted to full-wave simulation. The resulting responses are used to evaluate absorption performance and update the search history, guiding design improvements in the next iteration. AbsorbEvo also distils experience from training tasks into textual skills, which are independently validated before use in subsequent designs, without updating the base model weights. We develop AbsorbBench-36, a benchmark spanning explicit and implicit geometric representations, different material combinations and incidence conditions, to evaluate target attainment and search efficiency. Optimization trajectories and electromagnetic responses are analysed to examine design evolution, while independent skill transfer experiments assess the reuse of design experience across tasks.

## Results

*Overview of the AbsorbEvo framework*

AbsorbEvo translates natural-language performance objectives into absorber designs verified by full-wave simulations. Its dual-loop architecture coordinates three agents for planning and modelling, physics diagnosis and improvement, and experience evolution (Fig. 1). The agents share a frozen LLM and operate through role-specific instructions and tool configurations. The planning and modelling agent converts user requirements into a structured task specification and a simulation configuration, linking performance objectives and design constraints to executable modelling tools. Microwave-absorbing honeycomb sandwich structures (MAHSS) and triply periodic minimal surface (TPMS) structures represent explicit and implicit geometric descriptions, respectively.

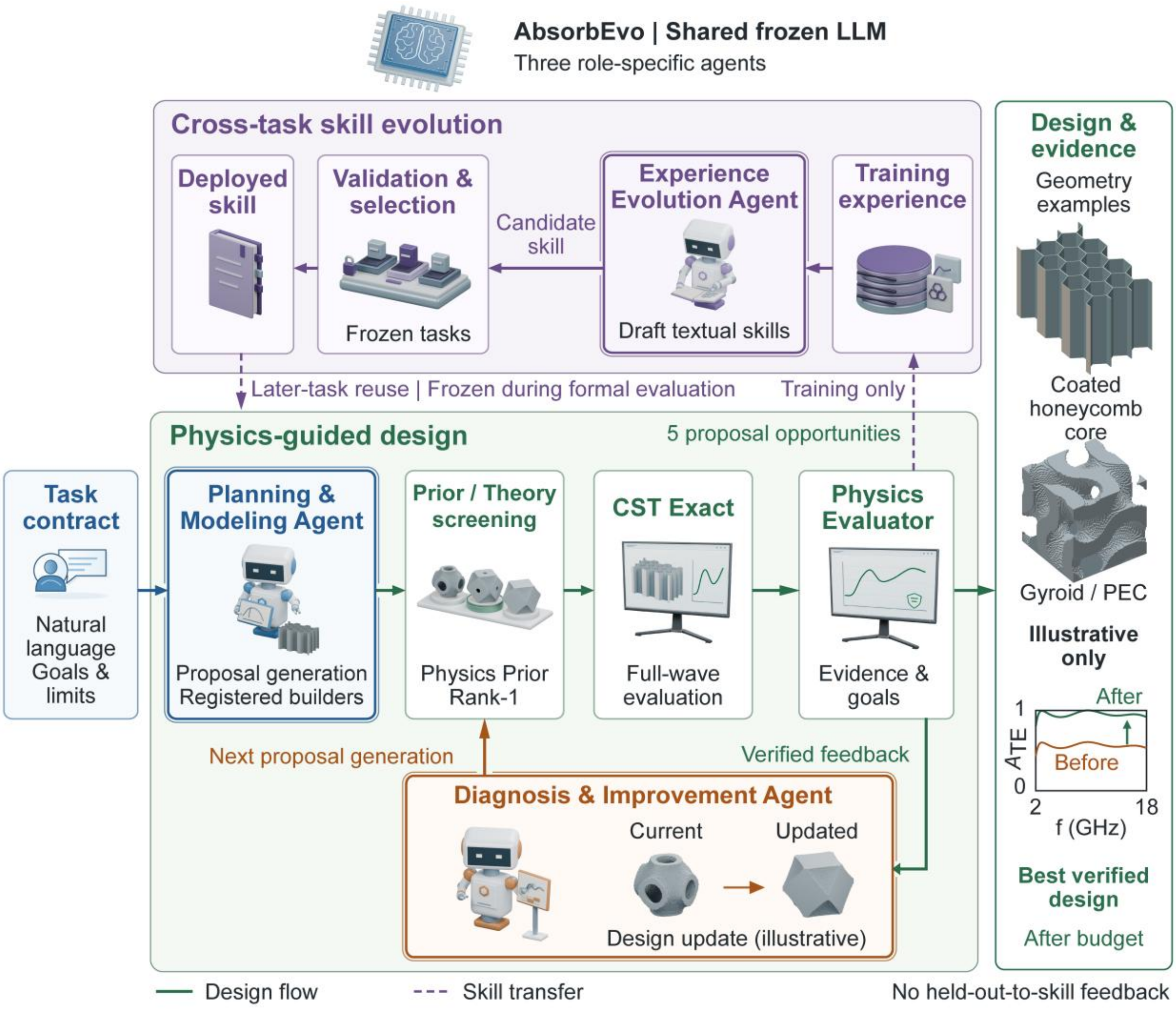

**Fig. 1 |** Dual-loop architecture of AbsorbEvo. The inner loop integrates task interpretation, modelling, physics verification and design improvement. The outer loop converts training experience into validated textual skills for subsequent tasks.

Within a task, language reasoning guides parameter adjustments, while physical predictions and full-wave results support candidate selection and subsequent design updates (Fig. 2). In the fixed-budget optimization setting, the language model proposes adjustment directions and magnitudes from the task objectives, reference design and search history. The system combines 36 directed candidates with 12 global candidates in each round. A low-cost physics prior ranks the candidates, and only the top-ranked design is submitted to full-wave simulation. Results that pass the applicable validity checks are added to the task history and used to guide the next round. Design quality is measured by coverage: the fraction of the target frequency band satisfying the specified reflection-loss threshold, averaged equally over the task-defined incidence conditions. Theoretical predictions support screening, whereas full-wave results determine target attainment and the final best design.

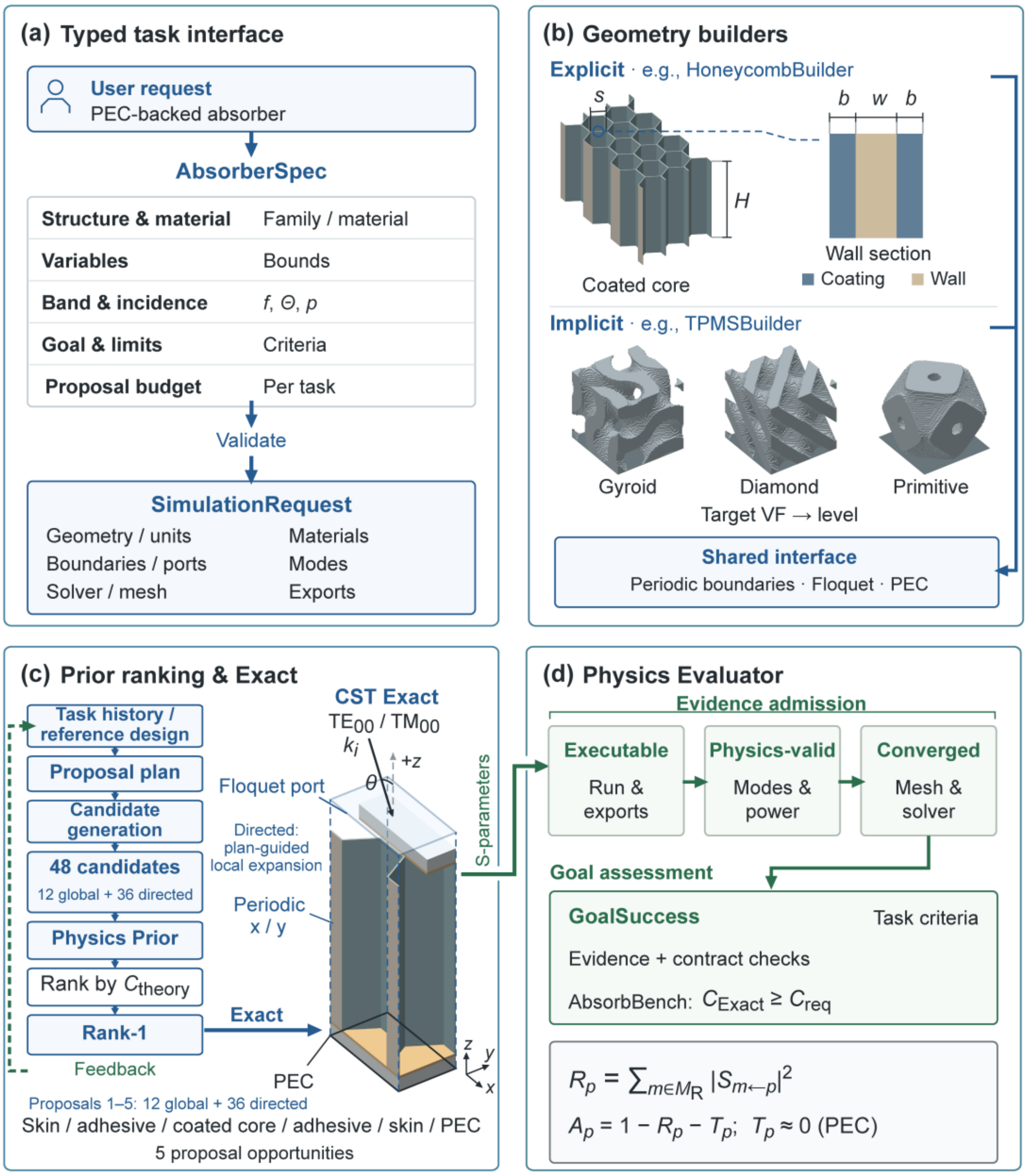

**Fig. 2** | Modelling and physics-guided evaluation. (a) Typed task interface. (b) Builders for MAHSS and TPMS structures. (c) Prior ranking and rank-1 full-wave evaluation. (d) Evidence admission and task-goal assessment.

The outer loop converts experience from training tasks into reusable textual guidance. The experience evolution agent distils design actions, computational results and error records into candidate skills, which are independently validated before use in subsequent tasks [15, 18]. This process updates external guidance rather than model weights. Guidance remains fixed during held-out evaluation. We assess the framework through system-level testing, fixed-budget optimization and an independent skill transfer experiment, distinguishing reliable task execution, within-task design improvement and cross-task experience reuse. Detailed implementations and evaluation protocols are provided in Methods and the Supplementary Information.

*Physics-prior agreement and system validation*

To assess the predictive agreement of the adopted effective-medium models, we performed paired comparisons of theoretical predictions and existing full-wave results for TPMS and MAHSS designs selected from previous simulations (**Fig. 3**a,b). For each case, both evaluations used identical geometric and material configurations, target frequency bands, reflection-loss thresholds and incidence conditions. Case selection and data processing are detailed in Supplementary Section S1.2.

For TPMS, the Spearman correlation coefficients for TE coverage, TM coverage and the mean coverage across both polarizations (Mean Joint) were 0.926, 0.893 and 0.935, respectively. The corresponding values for MAHSS were 0.980, 0.965 and 0.978. The mean and 90th-percentile absolute errors for both structural families are reported in **Table 1**.

Correlation measures rank agreement between the two evaluations, whereas absolute errors quantify deviations in predicted coverage. This comparison is limited to the selected designs and does not guarantee accurate performance predictions for every candidate. Theoretical predictions are therefore used for parameter screening, while design updates and target-attainment assessments remain based on full-wave results.

We next assessed whether AbsorbEvo could automate modelling and simulation and obtain valid results that met the design requirements. We compared single-shot inference (B0), execution retry (B1), and configurations using the evolved skill C1 without (B2) or with (B3) runtime evaluator feedback (Supplementary Table 5). All four configurations completed the simulations on the seven full-wave tasks, but B0 and B1 each failed the physical validity checks on one task. B2 and B3, which used the evolved skill, produced valid results that met the targets on all seven tasks (**Fig. 3**c). This difference suggests that validated experience-based guidance can improve the quality of task outcomes within this test set.

In one case (TE-X-T01), the simulation converged under the baseline configuration, but the reciprocity diagnostic value exceeded the predefined tolerance, so the result was not accepted as valid. In fault-detection testing, the evaluator detected all 20 predefined faults, including errors in complex power calculations and Floquet mode mapping, with no false positives among the 17 fault-free controls (**Fig. 3**d). These results demonstrate its ability to identify computational errors within this test set.

The five-task ablation study further examined the role of individual components. The full configuration successfully completed all five tasks. Removing range filtering or physics diagnosis, or reverting to the initial skill, reduced the number of successful tasks to three, four and three, respectively. The configuration without runtime evaluator feedback still completed all five tasks (**Fig. 3**c). For some configurations, the number of tasks with physically valid results exceeded the number of successful tasks, showing that passing physical validity checks does not imply target attainment.

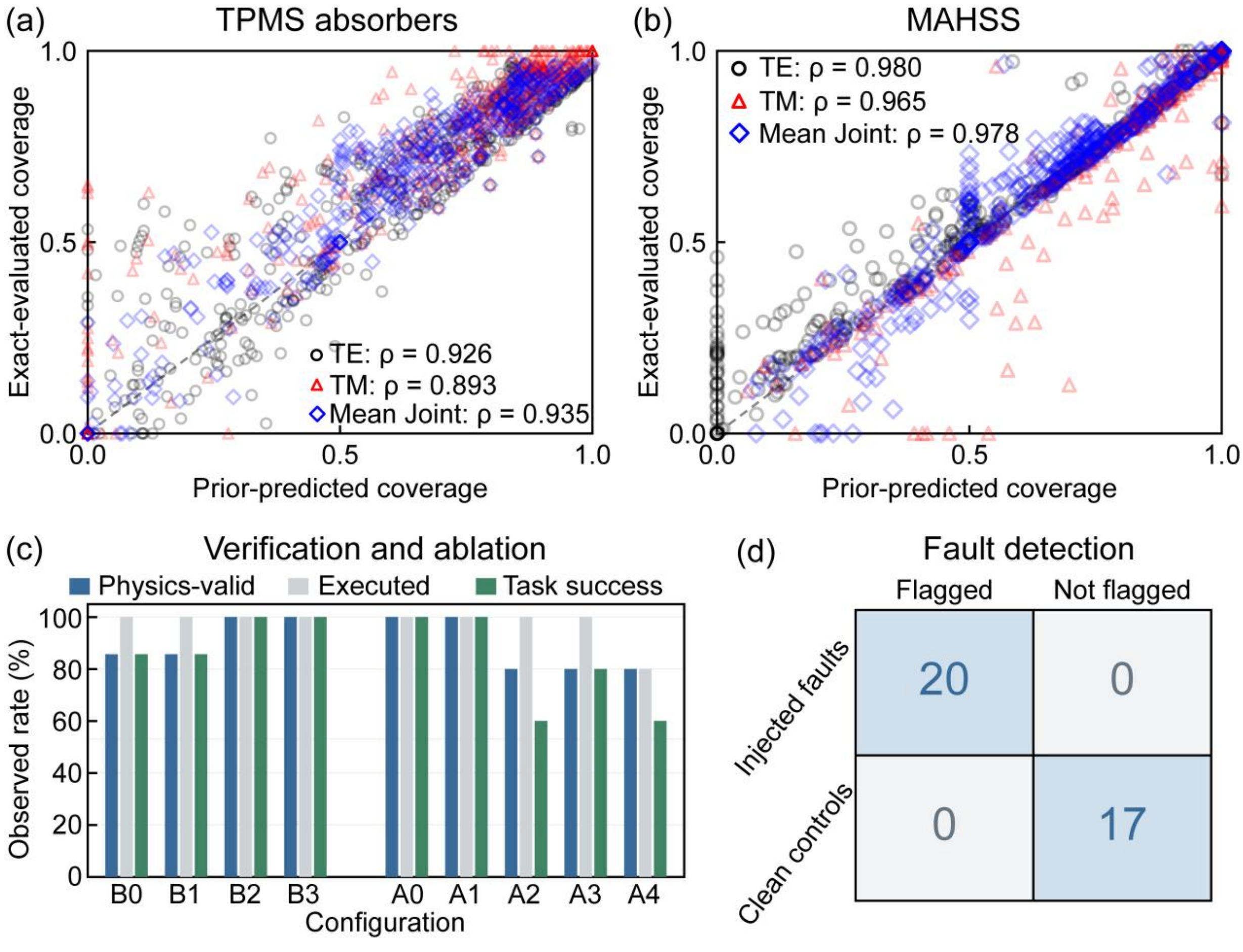

**Fig. 3** | Physics-prior agreement and system validation. (a,b) Prior–Exact coverage agreement for TPMS absorbers and MAHSS, respectively. (c) Execution, physical validity and task success on seven full-wave tasks (B0–B3) and five independent ablation tasks (A0–A4). (d) Detection outcomes for 20 injected faults and 17 clean controls.

**Table 1 |** Prior–Exact coverage agreement for TPMS and MAHSS.

| Structure | Response | Spearman ρ | MAE | P90 absolute error |
|---|---|---|---|---|
| TPMS (n = 474) | TE | 0.926 | 0.084 | 0.177 |
| | TM | 0.893 | 0.072 | 0.179 |
| | Mean Joint | 0.935 | 0.064 | 0.143 |
| MAHSS (n = 474) | TE | 0.980 | 0.040 | 0.121 |
| | TM | 0.965 | 0.029 | 0.056 |
| | Mean Joint | 0.978 | 0.033 | 0.096 |

n, the number of design–angle cases. Mean Joint is the equally weighted mean of TE and TM coverage.

*Target attainment and proposal efficiency under a fixed budget*

We compared AbsorbEvo with random search (Random), a generic agent (Generic Agent) and a configuration without new full-wave feedback (No-feedback) on 24 held-out tasks, comprising 12 MAHSS and 12 TPMS tasks. All methods received the same initial design for each task and five proposal opportunities. Detailed information-access rules and evaluation metrics are provided in Methods.

AbsorbEvo succeeded on 19 tasks (79.17%), compared with six (25.00%) for Generic Agent, 15 (62.50%) for No-feedback and three (12.50%) for Random (**Fig. 4**a and **Table 2**). AbsorbEvo's task success rate exceeded those of Generic and Random by 54.17 and 66.67 percentage points, respectively. All three language-model-based methods used gpt-6-luna with medium reasoning effort, allowing different design workflows to be compared using the same base model.

AbsorbEvo completed more tasks in fewer proposal rounds. By the third round, it had succeeded on 14 tasks, exceeding both Generic's two successes at the same stage and its six successes after all five rounds (**Fig. 4**d). Across the five TPMS tasks on which both methods succeeded, the mean first-success round was 2.60 for AbsorbEvo and 3.80 for Generic, corresponding to a reduction of 1.20 rounds (31.6%).

By structural family, AbsorbEvo succeeded on 9/12 MAHSS tasks, whereas neither Generic nor Random produced a design that met the target. For TPMS, AbsorbEvo and Generic succeeded on 10/12 and 6/12 tasks, respectively (**Fig. 4**b,c). This comparison suggests that, for the tested MAHSS tasks with coupled material and geometric parameters, a generic agent still struggles to meet performance targets within a limited number of rounds by proposing parameters directly from task requirements and previous feedback. By contrast, AbsorbEvo combines adjustment directions proposed by the language model with physics-based predictions, comparing multiple candidates before full-wave evaluation to provide an additional basis for selecting effective parameter combinations.

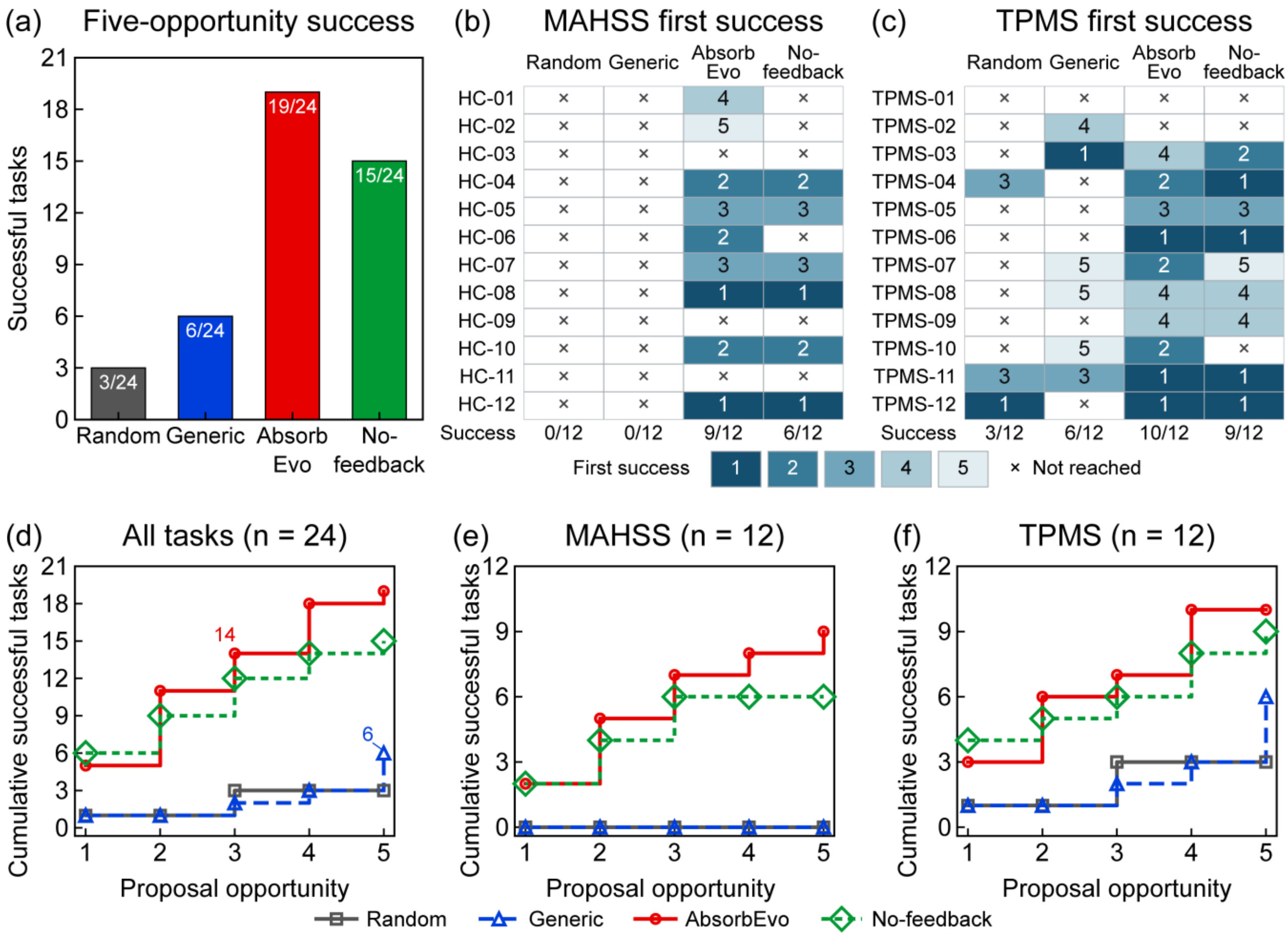

**Fig. 4** | Task attainment and proposal efficiency. (a) Successful tasks out of 24 within five proposal opportunities. (b, c) First-success opportunities for MAHSS and TPMS tasks; × denotes no success within budget. (d–f) Cumulative successful tasks for all tasks, MAHSS and TPMS, respectively.

To examine the role of search history, No-feedback excluded new full-wave feedback generated during optimization and used the shared initial design as its reference throughout, while retaining the same candidate-generation and physics-prior screening procedures. This configuration still succeeded on 15 tasks, outperforming Generic and Random and highlighting the contribution of physics-prior-guided candidate generation and screening. AbsorbEvo succeeded on all of these 15 tasks and four additional ones, indicating that using feedback from previous computations to refine subsequent designs helps more tasks reach their targets.

AbsorbEvo also achieved the highest mean best coverage, at 0.7816, compared with 0.6434 for Generic, 0.6448 for Random and 0.7613 for No-feedback. Of the 120 proposal opportunities allocated to each method, AbsorbEvo, Generic, Random and No-feedback produced 53, 13, five and 28 target-meeting designs, respectively, corresponding to 44.17%, 10.83%, 4.17% and 23.33% (Table 2). The four methods were allocated 480 proposal opportunities in total. Rounds that did not proceed to full-wave simulation because of non-compliant plans or insufficient candidates still counted towards the total budget, with coverage recorded as missing. Thus, in this comparison, AbsorbEvo's advantages were reflected in task success, mean best performance and the number of designs meeting the targets.

**Table 2 |** Four-method performance on 24 held-out tasks.

| **Method** | **Successful tasks, n (%)** | **Mean $C_5^{\mathrm{best}}$** | **Successful proposals** |
|---|---|---|---|
| Random | 3 (12.50) | 0.6448 | 5/120 |
| Generic Agent | 6 (25.00) | 0.6434 | 13/120 |
| **AbsorbEvo** | **19 (79.17)** | **0.7816** | **53/120** |
| No-feedback | 15 (62.50) | 0.7613 | 28/120 |

Five proposal opportunities per task and method.

*Physics-guided search and design evolution*

To visualize the search process, we constructed an independent two-dimensional MAHSS case by fixing the material parameters and retaining only coating thickness and core height as variables (**Fig. 5**). This case was not included in the 24-task held-out comparison. In each round, 12 global and 36 directed candidates were ranked together, and only the top-ranked design under the physics prior underwent full-wave evaluation. The global candidate selected in the first round increased coverage from 0.481 to 0.948. A directed candidate in the second round achieved 0.954, meeting the target of 0.95 for the first time; coverage increased further to 0.990 in the fourth round (**Fig. 5**b,c).

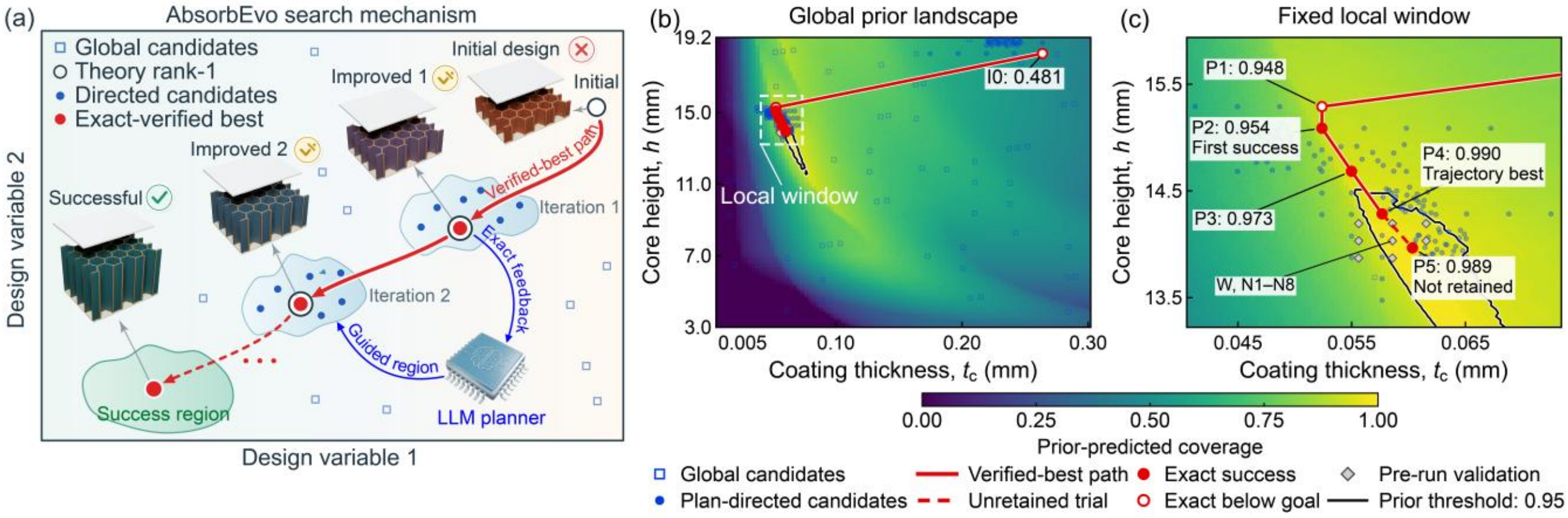

**Fig. 5 |** Physics-guided search and design evolution. (a) Conceptual search mechanism; structures and colours are illustrative. (b) Global prior-predicted coverage and the verified-best trajectory for a fixed-material MAHSS case. (c) Local view highlighting first success at P2, the best design at P4 and the unretained P5 trial. Black contours mark the prior-predicted 0.95 threshold; diamonds denote pre-run validation designs.

The prior-predicted coverage values for P2 and P3 were 0.910 and 0.940, respectively. Although these predictions were below the target, both designs met it in full-wave evaluation, showing that the prior-predicted threshold contour cannot be used directly as the actual target-attainment boundary. P5 had higher prior-predicted coverage than P4, but its full-wave coverage decreased slightly to 0.989, so P4 was retained as the best design. The setup of this independent two-dimensional case and the round-by-round candidate evolution are detailed in Supplementary Section S2.4 and Supplementary Fig. 1. This process illustrates AbsorbEvo's optimization strategy: global sampling broadens exploration, historical feedback guides directed improvements, and full-wave results determine which better-performing designs are retained, allowing subsequent search to build on existing valid designs.

*External skill evolution and transfer*

The independent skill transfer test assessed whether experience from training tasks could be converted into reusable guidance for absorber design. C1 was derived from the first batch of four training tasks. The selected skill C1 encoded experience in material referencing, Floquet mode checks and parameter adjustment as textual rules, retaining reusable procedures rather than case-specific optimal parameters.

On the six fixed validation tasks, the initial skill S0 succeeded on four, whereas C1 succeeded on all six and met the other evaluation requirements. Another candidate, C2, also succeeded on all six but did not outperform C1 in the overall evaluation. The proposed skill revision for C3 did not conform to the required format and was therefore excluded from numerical evaluation. C1 was thus selected for subsequent tasks based on independent validation rather than the language model's self-assessment. With identical attempt budgets and normalized computational costs, C1 succeeded on seven of the eight held-out tasks, compared with five for S0 (**Fig. 6**). Two previously unsuccessful tasks became successful, all five previous successes were retained, and the remaining task still failed to meet its target. Both additional successes occurred in the out-of-distribution group, increasing success from 1/3 to 3/3, while success on same-type and cross-family tasks remained at 3/3 and 1/2, respectively.

This test and the five-round AbsorbBench optimization comparison were conducted independently to assess cross-task experience reuse and within-task feedback, respectively. Skill content and model weights remained fixed throughout held-out testing.

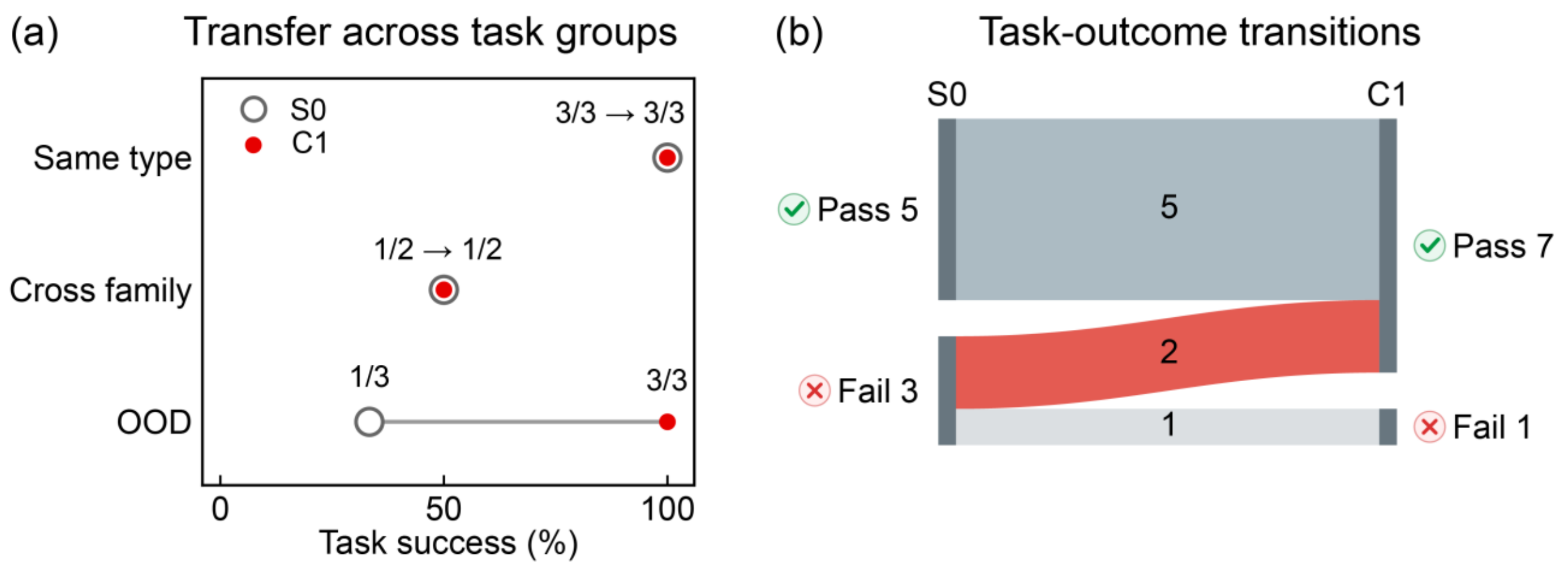

**Fig. 6** | Held-out transfer of evolved skills. (a) Task success with starter S0 and evolved C1 across same-type, cross-family and out-of-distribution tasks. (b) Outcome transitions for the eight held-out tasks; ribbon widths represent task counts.

*Representative physics-verified design cases*

Figure **7** links user requirements, AbsorbEvo's design plans and full-wave verification results for a MAHSS task and a TPMS task. The dialogues summarize the corresponding requirements and execution workflows.

For the MAHSS case (AB36-HC-T05), the requirement can be expressed as: "Design a MAHSS with an absorptance of at least 0.90 over at least 95% of the 2–18 GHz band under normal-incidence TE illumination." AbsorbEvo accordingly proposed a plan to adjust coating thickness, core height and material dispersion parameters within the task-defined bounds. The system generated directed and global candidates, ranked them using the physics prior to select a design for full-wave evaluation, and used the computational results to guide adjustments in the next round (**Fig. 7**a).

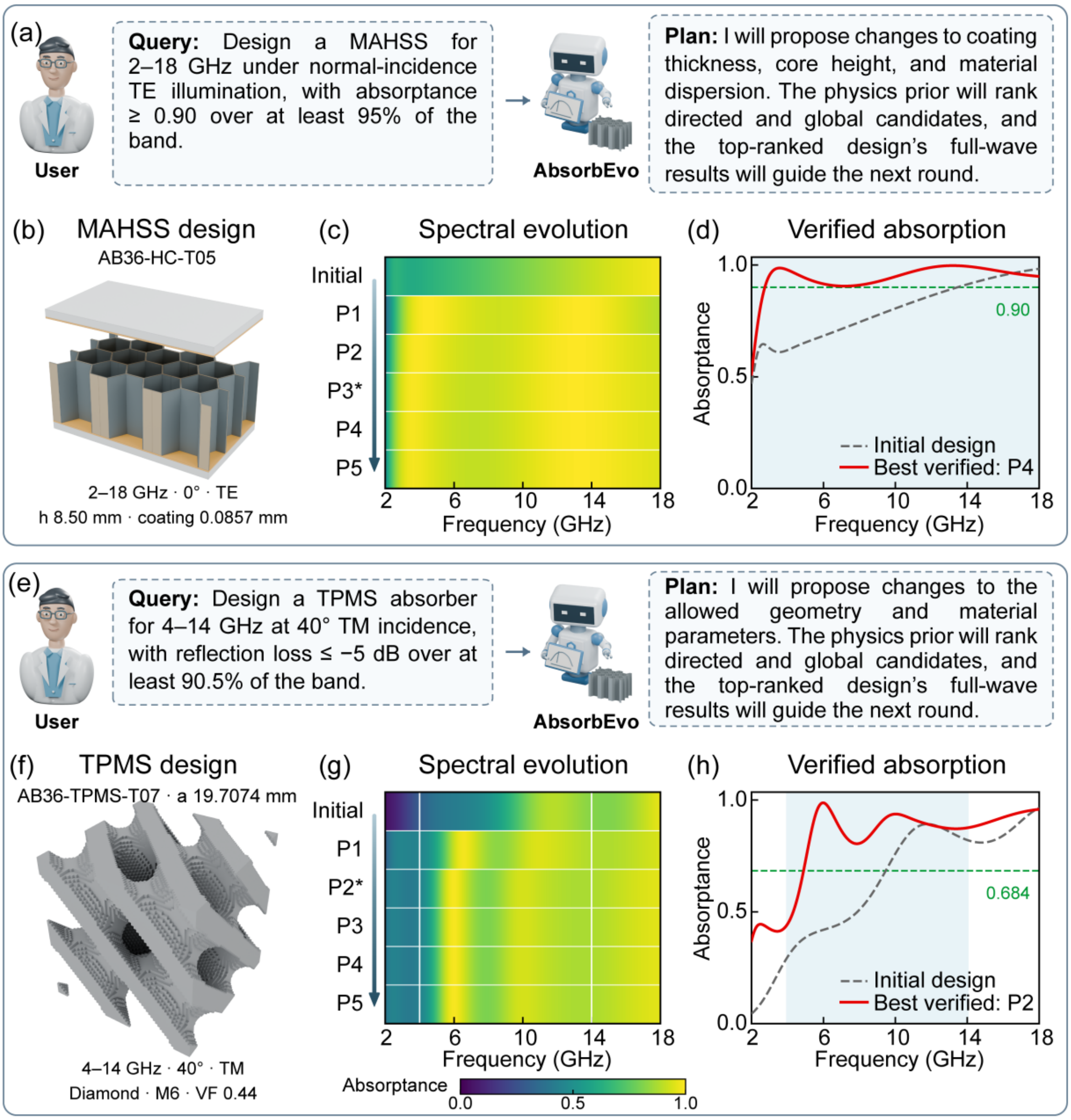

**Fig. 7** | Natural-language-driven absorber design. (a–d) MAHSS case AB36-HC-T05; (e–h) TPMS case AB36-TPMS-T07. Panels show schematic user queries and AbsorbEvo plans (a,e), best verified geometries (b,f), spectral evolution (c,g), and initial versus best verified spectra (d,h). P1–P5 denote proposal opportunities; asterisks mark first success. Shading and green dashed lines indicate target bands and absorptance thresholds, respectively.

Through iterative optimization, MAHSS coverage increased from 28.5% initially to 95.1% in the third round, meeting the target for the first time. It increased further to 95.3% in the fourth round, yielding the best design across all five rounds. Coating thickness decreased from 0.2638 to 0.0857 mm, and core height from 18.3022 to 8.5002 mm (**Fig. 7**b). Spectral evolution showed that the qualifying band extended towards lower frequencies, with the longest contiguous qualifying bandwidth increasing from 4.560 to 15.248 GHz (**Fig. 7**c,d). Coverage fell to 94.9% in the fifth round, so the fourth-round design was retained.

For the TPMS case (AB36-TPMS-T07), the requirement was: "Design a TPMS absorber with a reflection loss no greater than −5 dB over at least 90.5% of the 4–14 GHz band under TM-polarized illumination at an incidence angle of 40°." AbsorbEvo proposed adjustments to the permitted geometric and material parameters and followed the same prior-screening and full-wave-feedback workflow (**Fig. 7**e). The best design across the five rounds was obtained in the second round, increasing coverage from 45.24% to 90.95%. The design retained the Diamond topology but used M6 instead of M4. The target solid volume fraction increased from 0.29 to 0.44, and the period from 15.4130 to 19.7074 mm (**Fig. 7**f). The longest contiguous qualifying bandwidth increased from 4.524 to 9.095 GHz. All three subsequent rounds still met the target, but their coverage remained at 90.79%, below the second-round result (**Fig. 7**g,h).

These cases illustrate the design process from natural-language performance objectives through parameter adjustment, physics-prior screening and full-wave verification. Feedback from previous evaluations guides subsequent search, while final design selection is based on valid full-wave results, linking design adjustments to actual absorption performance.

## Discussion

AbsorbEvo integrates language reasoning, physics-based prediction and full-wave feedback to translate natural-language objectives into verified absorber designs. Rather than proposing parameters in isolation, the agent uses task objectives and search history to guide candidate generation, compares candidates through a physics prior and evaluates selected designs with full-wave simulations. Across 24 held-out tasks with five proposal opportunities per task, it achieved a task success rate of 79.17% and a mean best coverage of 0.7816, exceeding Generic Agent and No-feedback on both measures.

The comparisons highlight complementary roles for candidate screening and computational feedback. No-feedback retained the same candidate-generation and prior-screening procedures and still solved more tasks than Generic Agent and Random. Incorporating new full-wave feedback enabled AbsorbEvo to solve additional tasks. These results suggest that comparing candidate designs before expensive evaluation and using previous outcomes to refine subsequent adjustments can improve target attainment within a limited proposal budget. Independent physics evaluation also distinguishes valid computational results from designs that meet performance targets, helping prevent invalid responses from guiding later decisions. The agreement analysis and the two-dimensional example show why prior predictions support screening but do not replace full-wave verification.

The independent transfer experiment addresses a different form of adaptation: reuse of experience across tasks through external textual skills. Validated guidance improved success on new tasks while preserving all previously successful outcomes in the tested set. Because skill content and model weights remained fixed during held-out testing, this improvement reflects the use of previously selected guidance rather than updates driven by held-out results. The outer loop therefore complements within-task feedback without requiring changes to the base model weights.

The present evaluation used MAHSS and TPMS as representatives of explicit and implicit geometric descriptions. The results support the framework's effectiveness on the tested tasks, but its performance for other structures, materials and incidence conditions requires further validation. The low-cost prediction stage is not restricted to the effective-medium models used here; alternative predictive methods would require separate assessment of their accuracy and effects on optimization. Within this scope, combining physical knowledge, computational feedback and reusable experience provides a methodological basis for natural-language-driven autonomous absorber design.

## Methods

*Natural-language task interpretation and automated modelling*

The planning and modelling agent converts natural-language requirements into a structured task specification (AbsorberSpec), defining the geometric representation, material choices, adjustable parameters and their bounds, target frequency bands, incidence conditions and performance requirements. It then generates a simulation configuration (SimulationRequest), translating these requirements into geometry, material, boundary condition, excitation, mesh and solver settings to drive the corresponding modelling and computational tools. Geometry is modelled using explicit or implicit representations. Explicit representations define structures directly through dimensions, boundaries and relationships between components, whereas implicit representations use spatial functions and their isosurfaces to define structural form and the distribution of solid material. Both are implemented through corresponding parametric modelling modules, with a unified interface linking task specifications, geometry construction and performance evaluation (**Fig. 2**).

Design variables comprise geometric and material variables. The former control structural dimensions, form and solid material distribution, whereas the latter allow material selection or adjustment of dispersion parameters. Optimization updates these variables within task-defined bounds and uses the same geometric and material configuration for the physics prior and full-wave simulations, ensuring that predictions and performance evaluations consistently refer to the same design. Details of geometry construction, material parameters and full-wave simulation settings are provided in Supplementary Section S1.1.

*Deterministic physics verification*

Computational evaluation uses explicit numerical criteria and retains input and output records to support result verification [19, 20]. An independent physics evaluator verifies the geometry, material and excitation settings and checks output completeness, required modes, power relations and solver status. Only simulation results that pass these checks are accepted as valid evidence for performance evaluation and design updates. The physical validity criteria, numerical tolerances and their scope of applicability are detailed in Supplementary Section S1.3.

For frequency $f$, incidence angle $\theta$ and incident polarization $p$, the evaluator calculates reflectance, transmittance and absorptance from the scattering parameters:

$$R_p(f,\theta) = \sum_{m\in\mathcal{M}_{R,\text{prop}}} |S_{m\leftarrow p}|^2, T_p(f,\theta) = \sum_{n\in\mathcal{M}_{T,\text{prop}}} |S_{n\leftarrow p}|^2, A_p = 1 - R_p - T_p \tag{1}$$

Here, $\mathcal{M}_{R,\text{prop}}$ and $\mathcal{M}_{T,\text{prop}}$ denote the sets of propagating reflected and transmitted modes, respectively. The sums include both co- and cross-polarized components. Reflection loss is defined as

$$\text{RL}_p(f,\theta) = 10\log_{10} R_p(f,\theta) \tag{2}$$

With a perfect electric conductor (PEC) backing, transmittance is negligible, and absorptance is primarily determined by total reflectance. For an incidence condition $q = (\theta, p)$, the bandwidths of all intervals within the target band $[f_L, f_H]$ that satisfy the reflection-loss threshold $\tau_t$ are summed to obtain the total qualifying bandwidth $B_q$. The corresponding coverage is

$$C_q = \frac{B_q}{f_H - f_L}. \tag{3}$$

In numerical calculations, an interval contributes to $B_q$ only if both adjacent sampling points satisfy $RL_q \le \tau_t$. This bandwidth includes all qualifying frequency intervals, rather than only the longest contiguous band. For tasks involving multiple incidence angles or polarizations, coverage is averaged with equal weights:

$$C_t = \frac{1}{|Q_t|} \sum_{q \in Q_t} C_q, \tag{4}$$

where $Q_t$ is the task-specified set of incidence conditions. This metric represents the mean qualifying fraction of the target band across conditions, not the fraction satisfying all conditions simultaneously.

Coverage values predicted by the physics prior and obtained from full-wave simulations are denoted by $C_{\text{Theory}}$ and $C_{\text{Exact}}$, respectively. Both are calculated using Eqs. (3)–(4) with identical target frequency bands, reflection-loss thresholds and incidence conditions. The former enables rapid comparison of candidates, whereas the latter is used for performance evaluation and design updates. The physics prior model is described in Methods, "Design spaces and physics priors".

In AbsorbBench, a design is considered successful when it satisfies the predefined task constraints, its full-wave results pass the validity checks, and $C_{\text{Exact}} \ge C_{\text{req}}$. AbsorbEvo records the design parameters and valid results from each round in the task-specific search history, including attempts that do not meet the target. These records provide context for the physics diagnosis and improvement agent to assess the effects of previous adjustments and refine subsequent search directions. The modelling and physics evaluation workflow is shown in **Fig. 2**.

*Adaptive multi-fidelity optimization*

AbsorbEvo uses multi-fidelity optimization guided by a physics prior, screening parameter configurations through low-cost theoretical predictions and evaluating the absorption performance of selected designs through full-wave simulation. The results guide subsequent diagnosis and design adjustments, combining rapid parameter screening with high-fidelity evaluation. The overall optimization workflow is outlined in Supplementary Algorithm 1 (Supplementary Section S2.2).

In each round, the physics diagnosis and improvement agent identifies the variables to adjust and proposes the directions and magnitudes of changes based on task objectives, the reference design and search history. The system constructs parameter increments from this adjustment plan and adds them to the reference design to generate directed candidates. For continuous design parameters, this process is expressed as

$$\boldsymbol{x}_j^{\text{dir}} = \boldsymbol{x}^{\text{ref}} + \boldsymbol{\delta}_j. \tag{5}$$

Here, $\boldsymbol{x}^{\text{ref}}$ is the parameter vector of the reference design, $\boldsymbol{x}_j^{\text{dir}}$ is that of the $j$th directed candidate, and $\boldsymbol{\delta}_j$ is the corresponding parameter increment. The language model guides the directions and magnitudes of these increments through the adjustment plan, while the system generates their numerical values, allowing feedback from previous computations to inform subsequent designs. The system also samples global candidates across the entire design space to explore other promising designs and reduce the risk of becoming trapped in local optima. In this study, each round uses 36 directed candidates and 12 global candidates to form a pool of 48 designs after constraint checks and deduplication. The first round uses the shared initial design as its reference; subsequent rounds update the reference based on full-wave results.

The physics prior predicts the electromagnetic response of each candidate and ranks designs primarily by $C_{\text{Theory}}$, using the contiguous effective absorption bandwidth at the specified reflection-loss threshold as a secondary ranking criterion. In this study, only the top-ranked design undergoes full-wave simulation in each round to obtain responses across all target incidence angles and polarizations. Valid full-wave results and their corresponding design parameters from each round are added to the task-specific search history as context for diagnosis and adjustment in the next round. The reference for candidate generation is the design with the highest coverage among the task's valid new proposals; the shared initial design is used when no valid new proposal is available. The final best design is selected separately by comparing the initial design and all valid new proposals, with ties in coverage resolved in favour of the earlier design.

The number of optimization rounds can be set according to task requirements. For a consistent comparison across methods, this study uses five rounds per task, completing all rounds even if the target is reached earlier. The best valid design and the complete optimization record are returned. If an adjustment plan violates task requirements, or the prescribed number of unique, constraint-compliant candidates cannot be generated, full-wave simulation is skipped for that round, but the round still counts towards the total round budget.

*Design spaces and physics priors*

MAHSS and TPMS structures serve as examples of explicit and implicit geometric representations, respectively. MAHSS are described by cell dimensions, coating thickness, core height and the material properties of the absorbing coating; TPMS structures are described by topology, period, target solid volume fraction and material properties. Each task specifies adjustable variables, material choices and incidence conditions within the design spaces listed in **Table 3**.

**Table 3 |** Design spaces of the representative absorber structures.

| Representation and example | Design variables | Parameter ranges | Incidence conditions | Geometry and material settings |
|---|---|---|---|---|
| Explicit: MAHSS | Coating thickness; core height; material | Coating: 0.005–0.500 mm; core height: 3–30 mm; $\varepsilon_\infty$: 5–25; $\varepsilon_s$: 180–240; $\tau$: 0.05–0.10 ns | 0°, 30°, 60°; TE, TM or both | Coated 60° cell; task-fixed skins, adhesives and cell dimensions |
| Implicit: TPMS | Solid volume fraction; period; topology; material | Volume fraction: 0.10–0.90; period: 10–20 mm | 0°–70° in 10° increments; 45°; TE, TM or both | Primitive; Gyroid; Diamond Materials M2–M8 |

A low-cost physics prior based on effective-medium theory and wave propagation in layered media provides rapid predictions of reflection responses for parameter screening. The framework's low-cost prediction stage is not restricted to this implementation and can also incorporate pretrained and validated surrogate models, such as neural networks or neural operators. The numerical comparisons in this study use the theoretical models described below. The relative permittivity of the absorbing materials in both structural families is described by the Debye model:

$$\varepsilon_c(f) = \varepsilon_\infty + \frac{\varepsilon_s - \varepsilon_\infty}{1 + \mathrm{j}2\pi f\tau} \tag{6}$$

Here, $\varepsilon_\infty$ and $\varepsilon_s$ denote the high-frequency and static limits of the relative permittivity, respectively, and $\tau$ is the relaxation time. The time convention is $\exp(\mathrm{j}\omega t)$. The same material parameters are used for each candidate in the physics prior and full-wave simulations, ensuring a consistent material description across both evaluations.

The MAHSS model treats the core as an effective anisotropic medium while retaining the full layer stack. The TPMS model uses a single effective anisotropic layer constructed from the solid material distribution. Both models use a PEC backing. Details of effective-parameter calculation and volume-fraction mapping are provided in Supplementary Section S1.2. Once the effective permittivities are obtained, reflection responses are calculated using wave propagation relations for layered media. For an effective layer with permittivity tensor $\mathrm{diag}(\varepsilon_x, \varepsilon_y, \varepsilon_z)$, the propagation parameters and admittances for TE and TM polarizations are

$$\gamma_{\mathrm{TE}} = \sqrt{\varepsilon_y - \sin^2\theta}, Y_{\mathrm{TE}} = \gamma_{\mathrm{TE}}; \gamma_{\mathrm{TM}} = \sqrt{\varepsilon_x(1 - \sin^2\theta/\varepsilon_z)}, Y_{\mathrm{TM}} = \varepsilon_x/\gamma_{\mathrm{TM}} \quad (7)$$

Here, $\theta$ is the incidence angle, $\gamma$ is the normalized propagation parameter along the surface normal, and $Y$ is the normalized wave admittance for the corresponding polarization. Starting from $Z_{\mathrm{in},0} = 0$ at the PEC backing, the input impedance is recursively calculated layer by layer towards the incident surface. The final input impedance determines the reflection coefficient:

$$Z'_{\mathrm{in}} = Z_c \frac{Z_{\mathrm{in}} + Z_c \tanh(\beta d)}{Z_c + Z_{\mathrm{in}} \tanh(\beta d)}, Z_c = 1/Y, \beta = \mathrm{j}k_0\gamma, r = \frac{Z_{\mathrm{in}} - Z_0}{Z_{\mathrm{in}} + Z_0} \quad (8)$$

Here, $d$ is the thickness of the current layer, $k_0$ is the free-space wavenumber, and $Z_c$ is the layer's characteristic impedance. $Z_{\mathrm{in}}$ and $Z_{\mathrm{in}}'$ denote the input impedances before and after adding the current layer, respectively. When calculating $r$, $Z_{\mathrm{in}}$ is the final input impedance at the structure surface, and $Z_0$ is the wave impedance in air for the corresponding polarization. All impedances use the same normalization. This procedure yields the reflection response of the MAHSS layer stack.

For the single-layer TPMS model, the effective permittivity tensor is $\mathrm{diag}(\varepsilon_t, \varepsilon_t, \varepsilon_n)$, where $\varepsilon_t$ and $\varepsilon_n$ are the tangential and normal components, respectively. After calculating the propagation parameters and admittances using Eq. (7), the reflection coefficient can be written directly as

$$\rho = \frac{Y_0 - Y}{Y_0 + Y}, q = \exp(-2\mathrm{j}k_0 d\gamma), r = \frac{\rho - q}{1 - \rho q} \quad (9)$$

Here, $Y_0$ is the wave admittance in air for the corresponding polarization, $\rho$ is the reflection coefficient at the air–effective-medium interface, and $q$ is the round-trip propagation factor within the layer. For both structural families, the reflection coefficients yield the predicted reflection spectra. The target frequency bands, thresholds and incidence conditions specified in Methods, "Deterministic physics verification" are then used to calculate $C_{\mathrm{Theory}}$ for design comparison. Design updates and final performance evaluation remain based on $C_{\mathrm{Exact}}$ obtained from full-wave simulations.

*Experience accumulation and external skill evolution*

External textual skills convert task-specific computational experience into reusable guidance for subsequent designs. The experience evolution agent identifies recurring issues and effective strategies from design actions, valid responses and error records in training tasks, refining textual rules for material referencing, modelling, tool use, result checking and parameter adjustment. Successful outcomes help preserve effective practices, whereas results that fall short of targets and computational errors highlight areas for improvement.

Candidate and current skills are compared on the same set of validation tasks. Provided that physical validity, numerical quality and error-diagnosis capability do not deteriorate, target attainment, task success, performance metrics and computational cost are compared in a predefined order of priority. A candidate replaces the current version only if it performs better under these criteria; the existing version is retained in a tie. The selected skill is supplied to new tasks as external context, while model weights and other execution and evaluation settings remain unchanged during skill transfer evaluation. Training tasks are used to accumulate experience, validation tasks to select skills, and independent held-out tasks to assess transferability. Held-out results are not used to revise skills during testing.

*System-level evaluation*

System testing, fixed-budget optimization and skill transfer use distinct evaluation protocols to assess execution reliability, target attainment and experience reuse.

System-level evaluation assesses the execution of automated modelling and simulation, as well as the reliability of computational results and task-outcome assessments. The system test set comprises 24 tasks covering modelling, error diagnosis, design improvement and transfer. Eight held-out tasks are used to compare the four configurations listed in Supplementary Table 5 (Supplementary Section S2.5). For the seven full-wave tasks, execution, convergence, physical validity and target attainment are recorded separately. The remaining offline error-diagnosis task is scored for diagnostic correctness.

The evaluator's error-detection capability is tested using 20 predefined faults and 17 fault-free controls, with fault detections and false positives recorded separately. Single-component ablations are also performed on five independent tasks, using the full configuration as the reference. Each variant removes one component or replaces the skill version to assess its effect on task completion. The configurations are detailed in Supplementary Table 5 and scored by the same evaluator. These ablations use different tasks and protocols from the no-feedback optimization comparison in Methods, "Fixed-budget optimization", and their results are reported separately.

*Fixed-budget optimization*

AbsorbBench-36 comprises 18 MAHSS tasks and 18 TPMS tasks spanning different frequency bands, incidence angles, polarizations and parameter ranges. Each structural family contains four development tasks, two validation tasks and 12 held-out tasks, giving eight development, four validation and 24 held-out tasks in total. Method refinement used four preselected development tasks and all four validation tasks. The final comparison used the 24 held-out tasks. For each task, all methods received the same initial design and its full-wave evaluation results. The initial design passed the physical validity checks but did not meet the target coverage. Task objectives, design spaces and initial designs are detailed in Supplementary Sections S3.1–S3.2.

The methods compared are random search (Random), a generic agent (Generic Agent), AbsorbEvo and a configuration without new full-wave feedback (No-feedback). All methods use the same task conditions, modelling tools and evaluator, with five proposal opportunities per task. All five rounds are completed even if the target is reached earlier. In this fixed-budget comparison, Generic Agent, AbsorbEvo and No-feedback all use gpt-6-luna with the reasoning effort fixed at medium.

Random samples five candidates in a single batch before execution, without using a physics prior or computational feedback to adjust designs. Generic Agent directly proposes one candidate per round based on task requirements, the shared initial design and feedback from its own previous attempts within the current task. It does not use AbsorbEvo's physics-prior screening or external skills. AbsorbEvo uses the multi-fidelity search described in Methods, "Adaptive multi-fidelity optimization": the language model proposes an adjustment plan, the system generates candidates, and the physics prior ranks them to select a design for full-wave evaluation.

No-feedback retains the same language model, fixed domain guidance, candidate generation and physics-prior screening as AbsorbEvo, but does not use new full-wave feedback generated during the optimization run. Its reference remains the shared initial design throughout. Full-wave results are used only for recording and scoring, not for subsequent design adjustments, allowing the contribution of new feedback to the search process to be assessed. Domain guidance for both configurations remains fixed during testing, and skill transfer is evaluated separately as described in Methods, "Skill evolution and transfer". The inputs and feedback available to each method are detailed in Supplementary Section S2.1.

The primary metric is whether at least one design meets the target within five rounds. Let $i$, $m$ and $r$ denote the task, method and proposal round, respectively. The round-level success indicator is $g_{i,m,r} = 1$ if the candidate satisfies the task constraints, its full-wave results pass the validity checks in Methods, "Deterministic physics verification", and its coverage reaches the target $C_{\mathrm{req},i}$; otherwise, $g_{i,m,r} = 0$. The five-round target-attainment indicator is defined as

$$\mathrm{GS}_{5,i,m} = \max_{1\le r\le 5} g_{i,m,r}. \tag{10}$$

The task success rate is the proportion of successful tasks among all test tasks. Rounds without valid full-wave results still count towards the proposal budget, and their coverage is recorded as missing. In addition to task success rate, best valid coverage and the first-success round are used to assess design quality and search efficiency. Let $C_{0,i}$ denote the initial coverage of task $i$, $C_{i,m,r}$ the valid full-wave coverage obtained by method $m$ in round $r$, and $\mathcal{V}_{i,m,k}$ the set of rounds among the first $k$ with valid results and constraint-compliant designs. The best coverage up to round $k$ is then

$$C_{i,m,k}^{\mathrm{best}} = \max\left(\{C_{0,i}\} \cup \{C_{i,m,r}: r \in \mathcal{V}_{i,m,k}\}\right),\ k = 1, \ldots, 5. \tag{11}$$

The first-success round is defined as

$$T_{i,m} = \min\{r \in \{1, \ldots, 5\}: g_{i,m,r} = 1\}. \tag{12}$$

Tasks that do not succeed within five rounds are labelled "target not reached". For a test set containing $N$ tasks, the cumulative proportion of successful tasks up to round $k$ is

$$F_m(k) = \frac{1}{N}\sum_{i=1}^{N} \max_{1\le r\le k} g_{i,m,r},\ k = 1, \ldots, 5. \tag{13}$$

These metrics respectively quantify the best performance of the designs obtained, the number of rounds required to reach the target, and the proportion of successful tasks as optimization progresses. Between-method comparisons are paired by task. First-success rounds are compared only for tasks on which both methods succeed, while cumulative success counts across all tasks characterize overall search performance.

*Skill evolution and transfer*

This test assesses whether design experience helps the agent complete new tasks. Ten training tasks were executed in three batches. After each batch, experience from completed tasks was summarized to revise the textual skill. Each candidate skill was compared with the current version on the same six validation tasks and adopted only if it performed better. The selected version was retained for held-out testing.

The agent was then evaluated on the same eight held-out tasks using the initial skill S0 and the selected skill C1 in separate runs. The tasks were grouped into same-type, cross-family and out-of-distribution categories. Both runs used the same model, attempt budget and evaluation settings, with no skill updates during testing. Task success rates were compared, and failure-to-success and success-to-failure transitions were recorded to assess whether the new skill enabled more tasks to be solved while preserving existing capabilities. This test was conducted independently of the five-round AbsorbBench optimization comparison. Task settings, skill content and selection, and cost calculation rules are detailed in Supplementary Sections S4.1, S4.2 and S4.3, respectively.

## Data availability

The AbsorbBench-36 benchmark is available at https://github.com/ZhichengFeng/AbsorbEvo/tree/main/benchmarks/AbsorbBench-36.

## Code availability

The complete source code for this study will be made publicly available upon acceptance of the manuscript.

## Author contributions

Zhicheng Feng: Investigation, Data curation, Methodology, Software, Conceptualization, Writing – original draft. Yubo Zhao: Validation, Data curation. Xuefeng Yao: Writing – review & editing, Funding acquisition, Supervision.

## Funding

This work was supported by the National Science and Technology Major Project of China under Grant No. 2025ZD1603600.

## Competing interests

The authors declare that they have no known competing financial interests or personal relationships that could have appeared to influence the work reported in this paper.

# Supplementary Information

AbsorbEvo: An Agentic Framework for Autonomous Inverse Design of Microwave Absorbers

Zhicheng Feng[1], Yubo Zhao[1], Xuefeng Yao[1]*

[1]*Department of Engineering Mechanics, Applied Mechanics Lab, Tsinghua University, Beijing 100084, P. R. China*

*Corresponding author. Email: yxf@mail.tsinghua.edu.cn

## Contents

This supplementary information document includes the following sections:

## S1. Electromagnetic models and physical verification

### S1.1. Geometry, materials and full-wave settings

Microwave-absorbing honeycomb sandwich structures (MAHSS) comprise skins and adhesive layers on both sides of a coated honeycomb core, with a perfect electric conductor (PEC) back boundary. Here, h is the core height, s the hexagonal cell-wall side length, w the wall thickness and b the one-sided absorbing coating thickness. The cell angle $\alpha = 60°$ defines a regular hexagon. Geometric ranges and material parameters are given in Supplementary Table 7 and Supplementary Table 2, respectively.

Triply periodic minimal surface (TPMS) structures include Primitive, Gyroid and Diamond topologies. For a period a, the normalized coordinates are $X = \frac{2\pi x}{a}, Y = \frac{2\pi y}{a}, Z = \frac{2\pi z}{a}$. Their implicit functions are, respectively,

$$\varphi_{\mathrm{P}} = \cos X + \cos Y + \cos Z \tag{S1}$$

$$\varphi_{\mathrm{G}} = \sin X \cos Y + \sin Y \cos Z + \sin Z \cos X \tag{S2}$$

$$\begin{aligned} \varphi_{\mathrm{D}} = {} & \sin X \sin Y \sin Z + \sin X \cos Y \cos Z \\ & + \cos X \sin Y \cos Z + \cos X \cos Y \sin Z \end{aligned} \tag{S3}$$

The TPMS solid region is defined by a threshold band of the implicit function and discretized using $64^3$ voxel centres per period. Either the threshold half-width or the target solid volume fraction (VF) determines the solid region. Parameter ranges are given in Supplementary Table 7.

**Supplementary Table 1 | Geometry and full-wave settings.**

| Setting | MAHSS | TPMS |
|---|---|---|
| Layer stack | Skin / adhesive / coated honeycomb core / adhesive / skin / PEC | Air / TPMS dielectric structure / PEC |
| Thickness | Skin: 1.000 mm; adhesive: 0.200 mm (each) | a (one period) |
| Cell geometry | Wall side s; wall thickness w; angle 60° | Primitive, Gyroid or Diamond |
| Solid region | Coating b on each wall face | $\lvert\varphi - c_0\rvert \le t$; $c_0 = 0$ |
| Discretization | Extruded profiles | $64^3$ voxel centres; 1 × 1 × 1 cells |
| Side boundaries | Periodic in x and y | Periodic in x and y |
| Port / air gap | Zmax Floquet / 1.250 mm | Zmax Floquet / a/8 |
| Incidence | Task-specific θ; φ = 0° | Task-specific θ; φ = 0° |
| Solver | CST 2025; frequency-domain FEM | CST 2025; frequency-domain FEM |
| Band / output grid | 2–18 GHz / 1001 points | 2–18 GHz / 1009 points |
| Frequency sampling | Auto interval 50; ≥15 broadband samples | Auto interval 50; ≥15 broadband samples |
| Modes | TE00/TM00; co/cross reflection | TE00/TM00; co/cross reflection |
| Mesh | 2nd-order tetrahedra; near/far 17/1; nonadaptive | 2nd-order tetrahedra; near/far 17/1; nonadaptive |
| Accuracy setting | AccuracyTet = $10^{-4}$ | AccuracyTet = $10^{-3}$ |

Full-wave calculations use the frequency-domain finite-element method (FEM) solver in CST Studio Suite 2025, with periodic boundaries in x and y and a Zmax Floquet port. The incidence angle θ is task-specific, whereas the azimuth angle is fixed at φ = 0°. Transverse electric (TE) and transverse magnetic (TM) polarizations are excited separately for the same incident direction. Both polarizations are computed over 2–18 GHz, with coverage evaluated within the target band of each task. Other settings are listed in Supplementary Table 1.

**Supplementary Table** 2 **| Material models and parameters.**

| Material | Model | Parameters |
|---|---|---|
| Nomex wall | Lossy dielectric | εr = 3.800; tan δ = 0.015 |
| Skin | Lossy dielectric | εr = 3.300; tan δ = 0.015 |
| Adhesive | Lossy dielectric | εr = 3.400; tan δ = 0.015 |
| Coating | Debye | $\varepsilon_\infty \in [5, 25]$; $\varepsilon_s \in [180, 240]$; τ ∈ [0.050, 0.100] ns |
| M1 (skill tasks) | Debye | $\varepsilon_\infty = 2.535$; $\varepsilon_s = 2.838$; τ = 0.006 ns |
| M2 | Debye | $\varepsilon_\infty = 3.003$; $\varepsilon_s = 3.751$; τ = 0.008 ns |
| M3 | Debye | $\varepsilon_\infty = 3.686$; $\varepsilon_s = 5.240$; τ = 0.010 ns |
| M4 | Debye | $\varepsilon_\infty = 4.408$; $\varepsilon_s = 6.905$; τ = 0.012 ns |
| M5 | Debye | $\varepsilon_\infty = 5.154$; $\varepsilon_s = 8.783$; τ = 0.014 ns |
| M6 | Debye | $\varepsilon_\infty = 6.014$; $\varepsilon_s = 11.214$; τ = 0.016 ns |
| M7 | Debye | $\varepsilon_\infty = 6.905$; $\varepsilon_s = 13.958$; τ = 0.018 ns |
| M8 | Debye | $\varepsilon_\infty = 7.760$; $\varepsilon_s = 17.406$; τ = 0.020 ns |

All dielectrics have μr = 1. Debye models are single-pole, with exp(+jωt) and τ in ns.

## S1.2. Implementations of the physics priors

The physics priors combine effective-medium and layered-propagation models. For TPMS, q denotes a voxel centre, $c_0 = 0$ the threshold centre, and t the threshold half-width. The solid indicator function and target volume-fraction mapping are

$$\chi_t(\mathbf{q}) = \mathbf{1}\{|\varphi(\mathbf{q}) - c_0| \le t\}$$
$$t^* = \arg\min_{t\in\mathcal{T}} \left|\frac{1}{N_\mathrm{v}}\sum_{\mathbf{q}}\chi_t(\mathbf{q}) - v_\mathrm{target}\right| \quad \text{(S4)}$$

Here, $N_\mathrm{v} = 64^3$, and T contains the discrete values of $|\varphi - c_0|$ at the voxel centres.

The MAHSS prior retains the skins and adhesive layers and represents the coated honeycomb core as an effective layer with a diagonal permittivity tensor. Let $\varepsilon_{\mathrm{N}}$, $\varepsilon_{\mathrm{C}}$ and $\varepsilon_{\mathrm{A}}$ denote the relative permittivities of the wall, coating and air, respectively, with $\ell = s\sin\alpha, V = s^2\sin\alpha\cos\alpha$. At α = 60°, the material fractions and parallel and series equivalents in the two regions are

$$\mathbf{u}_1 = \frac{(w,b,\ell-w-b)}{\ell}, \mathbf{u}_2 = \frac{(ws,2bs,V-ws-2bs)}{V}$$
$$P_i = \sum_{m\in\{\mathrm{N,C,A}\}} u_{i,m}\varepsilon_m, H_i = \left(\sum_{m\in\{\mathrm{N,C,A}\}} \frac{u_{i,m}}{\varepsilon_m}\right)^{-1} \quad \text{(S5)}$$

The end-wall and inclined-wall regions have area weights of 2/3 and 1/3, respectively, giving the core permittivity tensor

$$\varepsilon_x = \left[\frac{2}{3P_1} + \frac{1}{3\left(\frac{3H_2}{4}+\frac{P_2}{4}\right)}\right]^{-1}$$
$$\varepsilon_y = \frac{2H_1}{3} + \frac{\frac{H_2}{4}+\frac{3P_2}{4}}{3}, \varepsilon_z = \frac{2P_1+P_2}{3} \quad \text{(S6)}$$

Both regions require a positive air fraction, with $\varepsilon_{\mathrm{N}}$ = 3.800(1 − j0.015) and $\varepsilon_{\mathrm{A}}$ = 1.001. TE propagation uses $\varepsilon_y$, whereas TM propagation uses $\varepsilon_x$ and $\varepsilon_z$. The input impedance of the five layers is calculated recursively from the PEC boundary towards the incident side. The prior neglects cross-polarized coupling, whereas full-wave evaluation includes both co-polarized and cross-polarized reflected power.

The TPMS structure is divided into N equal-thickness slices along z. Let $c_l$ denote the solid area fraction in slice l, and $\varepsilon_{\mathrm{solid}}$(f) the complex permittivity of the solid. With background permittivity $\varepsilon_{\mathrm{b}}$ = 1, the effective slice properties and their combination across layers are

$$\varepsilon_{l,a} = \varepsilon_{\mathrm{b}} + \frac{c_l(\varepsilon_{\mathrm{solid}}-\varepsilon_{\mathrm{b}})}{1+(1-c_l)L_a\frac{(\varepsilon_{\mathrm{solid}}-\varepsilon_{\mathrm{b}})}{\varepsilon_{\mathrm{b}}}}$$
$$(L_x, L_y, L_z) = (0,0,1), \beta_l = \frac{1}{N} \quad \text{(S7)}$$
$$\varepsilon_{x,y} = \sum_l \beta_l \varepsilon_{l,x,y}, \varepsilon_z = \left(\sum_l \frac{\beta_l}{\varepsilon_{l,z}}\right)^{-1}$$

Equation (S7) reduces to $\varepsilon_x = \varepsilon_y = 1 + v(\varepsilon_{\mathrm{solid}} - 1), \varepsilon_z = \left(1 - v + \frac{v}{\varepsilon_{\mathrm{solid}}}\right)^{-1}$, where v is the mean solid fraction across the slices. This approximation therefore gives the same effective response for identical materials, thickness and v. Prior ranks candidates, whereas Exact provides full-wave evaluation; both use the same materials, geometry and task conditions.

main-text Fig. 3 and Table 1 in the main text reuse historical full-wave results from the R17 development/validation comparison, the R18 held-out comparison and their registered Initial designs. Cases were deduplicated by actual geometry, material, incidence angle, target band and threshold. All 474 MAHSS cases were included. For TPMS, 474 of 531 eligible records were selected by stratifying by source and task, allocating proportional quotas, and ranking case hashes with a fixed seed. Records were excluded if results were incomplete or invalid, channel exports were incorrect, or conditions did not match the formal objectives. Selection did not depend on coverage, prediction error or task success. Prior and Exact used identical task conditions and frequency grids. Mean Joint is the equally weighted mean of TE and TM coverage for the same case. Coverage was retained to 12 decimal places when handling numerical ties. No additional full-wave simulations were performed, and the agreement metrics apply only to the selected cases.

**S1.3. Physical verification criteria**

Physical validity checks determine whether a result can be used for design evaluation; target attainment is assessed separately against the task objectives. The criteria are summarized in Supplementary Table 3. The two-mode reflection spectra must contain four finite complex co-polarized and cross-polarized reflection coefficients for TE00 and TM00 incidence, consistent with the task inputs. The total reflected power for both polarizations satisfies

$$R_p(f_k) = \left|r_{\mathrm{co},p}(f_k)\right|^2 + \left|r_{\mathrm{cross},p}(f_k)\right|^2 \\ \max_{k,p} R_p(f_k) \le 1 + \varepsilon_{\mathrm{P}}, \varepsilon_{\mathrm{P}} = 5 \times 10^{-3} \tag{S8}$$

For PEC backing, T = 0 and A = 1 − R. The power-reconstruction and energy-consistency residuals are defined as

$$e_{\mathrm{R}} = \max_{k,p}\left|R_p^{\mathrm{export}}(f_k) - \hat{R}_p(f_k)\right| \\ e_{\mathrm{E}} = \max_{k,p}\left|R_p^{\mathrm{export}}(f_k) + T_p^{\mathrm{export}}(f_k) + A_p^{\mathrm{export}}(f_k) - 1\right| \tag{S9}$$

The superscript export denotes exported power spectra, and reflected power is reconstructed using Eq. (S8). Both residual thresholds are $10^{-10}$; when A = 1 − R, $e_{\mathrm{E}}$ measures post-processing consistency. The complex amplitude difference between the cross-polarized channels is

$$e_{\mathrm{rec}} = \max_{k}\left|r_{\mathrm{TE}\leftarrow\mathrm{TM}}(f_k) - r_{\mathrm{TM}\leftarrow\mathrm{TE}}(f_k)\right| \tag{S10}$$

This residual includes both magnitude and phase differences, with a threshold of $10^{-4}$. It diagnoses channel consistency under the specified port, mode and incidence conditions.

The fixed-budget comparison checked input consistency, solver and export status, the registered two-mode reflection spectra, and the reflected-power upper bound for each case. Power reconstruction, energy consistency, cross-channel differences, and associated log and mesh diagnostics were used only in historical system-validation configurations where they were enabled. They were not independently checked for every result in the fixed-budget comparison.

When absorptance is obtained from $A = 1 - R$, the power-reconstruction and energy-consistency residuals check export and post-processing consistency. They do not constitute an independent energy-closure test based on field integration. Geometry and material checks are limited to build records. Checking the two fundamental modes does not imply independent verification of all higher-order propagating modes. Coating-sheet warnings for the coated honeycomb core are retained.

**Supplementary Table** 3 **| Physical verification criteria.**

| Check | Quantity / evidence | Criterion |
|---|---|---|
| **Checks used in the fixed-budget comparison** | | |
| Input consistency | Geometry, material, angle, grid and outputs | Matching inputs; complete data |
| Mode spectra | Four complex TE00/TM00 reflection terms | Complete, finite co/cross spectra |
| Reflected power | Co/cross power sum, Eq. (S8) | max R ≤ 1.005 |
| Solver completion | Solver log and spectra | Successful termination; complete output |
| Geometry / material | Builder inputs and receipts | Input consistency; coating-sheet warning retained |
| **Additional diagnostics in the applicable historical configurations** | | |
| Power reconstruction | $e_{\mathrm{R}}$, Eq. (S9) | $e_{\mathrm{R}} \leq 10^{-10}$ |
| Energy consistency | $e_{\mathrm{E}}$, Eq. (S9) | $e_{\mathrm{E}} \leq 10^{-10}$ |
| Cross-channel difference | Complex amplitude difference, Eq. (S10) | $e_{\mathrm{rec}} \leq 10^{-4}$ |
| Solver accuracy | Model.log accuracy and registered sample count | Accuracy ≤ $10^{-3}$; samples ≥ 15 |
| Mesh comparison | Coarse/fine results on a common grid | Max ΔRL ≤ 0.100 dB; fine cells ≥ 1.250 × coarse |
| Warning thresholds | Failed finite residuals; later evaluator | $e_{\mathrm{R}} \leq 10^{-6}$; $e_{\mathrm{E}} \leq 10^{-6}$; $e_{\mathrm{rec}} \leq 10^{-3}$; solver-log residual ≤ $10^{-2}$ |

Notes. The two groups apply to different evaluation settings. Additional diagnostics were used only where enabled in the corresponding historical configuration, not as independent checks of every fixed-budget result. Warning thresholds classify existing failed checks; they do not replace the nominal criteria. Input-record and two-mode checks do not constitute independent validation of all geometry details or higher-order propagating modes.

## S2. Agent implementation and search settings

### S2.1. Information access

Shared inputs to the large language model (LLM) include task requirements, the design domain, initial parameters and coverage $C_0$. Subsequent information is supplied according to each method. Performance feedback comprises summary results such as coverage, without frequency-resolved reflectance data. The four methods differ in candidate generation and their use of the physics prior and full-wave feedback (Supplementary Table 4). Random samples designs randomly, whereas Generic Agent directly proposes design parameters. AbsorbEvo uses fixed physics guidance to formulate a search plan, selects a candidate through Prior ranking, verifies it with Exact, and updates the plan from the results. No-feedback uses the same guidance and selection procedure, but retains only previous parameters and plans for subsequent decisions, without receiving new Exact results.

Fixed domain guidance was established before the formal comparison and remained unchanged throughout the five rounds. It covers how MAHSS geometry and Debye parameters affect matching and loss, adjustments to TPMS geometry and materials, and conservative diagnosis based on the supplied responses. AbsorbEvo and No-feedback use the same guidance; Generic Agent does not receive this dedicated guidance.

The three LLM-based methods in the formal fixed-budget comparison used gpt-6-luna with medium reasoning effort. Random does not call an LLM. Each method uses only information permitted for the current task and its own records, without accessing solutions from other tasks or methods. No-feedback receives no new Exact responses from the current optimization. Historical system verification and the independent skill evaluation used gpt-5.6-terra, whereas the two-variable example used gpt-6-astra, both with medium reasoning effort.

**Supplementary Table 4 | Information and candidate selection for the four methods.**

| Method | Input | Exact feedback | Physics guidance / Prior | Candidate selection |
|---|---|---|---|---|
| Random | Design domain; shared Initial | Unused | None | Five random candidates fixed before evaluation |
| Generic Agent | Task, Initial/$C_0$, own history | Coverage, validity and success | None | One model-proposed design per round |
| AbsorbEvo | Task, Initial/$C_0$, own history | Coverage, validity, compliance and success | Fixed guidance and Prior | 36 directed + 12 global; Prior rank 1 |
| No-feedback | Task, Initial/$C_0$, past parameters/plans | Withheld | Same as AbsorbEvo | 36 directed + 12 global; Initial reference |

### S2.2. Candidate generation and selection

In the formal fixed-budget comparison, AbsorbEvo and No-feedback generate 36 directed and 12 global candidates per round. The LLM selects increase, decrease or hold for each variable, a normalized step of 0.050, 0.100 or 0.200, and an overall scale of 0.500, 1.000 or 1.500. At least two variables must be enabled, with at least one assigned a nonzero adjustment direction. Candidate counts and the number of rounds are fixed in advance.

Continuous variables are normalized to their design ranges, and categorical variables are mapped to equally spaced indices. Directed search includes single-variable, two-variable and multivariable adjustments, with single-variable step multipliers of 0.500, 1, 1.500 and 2. Remaining slots are filled by local samples distributed according to fixed rules using a Halton sequence. Enabled variables assigned hold are sampled on both sides of the reference point. Other enabled variables are sampled along the specified direction at 0.350–1.000 times the step. Variables that are not enabled retain their reference values.

Global candidates use the same sampling method across the full design domain. AbsorbEvo and No-feedback use identical seeds for corresponding tasks and rounds, and the 12 global candidates are independent of the LLM plan. Out-of-range and duplicate candidates are discarded; duplicate checks cover the current pool, the reference point and all previous candidate pools. TPMS VF values lie on a 0.010 grid, and categorical variables are mapped to the nearest permitted category. Prior ranking begins only after the complete 36+12 candidate pool is formed. Valid, applicable predictions are prioritized, followed by descending prior-predicted coverage and then descending continuous effective absorption bandwidth (EAB). Remaining ties favour, in order, a smaller response shortfall relative to the threshold, a larger normalized distance from the generation reference, and a smaller candidate index. Only the top-ranked design is submitted to Exact in each round.

AbsorbEvo uses the design with the highest coverage among valid new proposals as the reference for candidate generation. The shared initial design is used only when no valid new proposal is available. The final best design is selected separately from the initial design and all valid new proposals, retaining the earlier design in a tie. No-feedback always uses the shared initial design as its generation reference.

Supplementary Algorithm 1 summarizes the physics-guided search, using the candidate-generation, ranking and reference-update rules described above.

**Supplementary Algorithm 1 | Physics-guided multi-fidelity search with AbsorbEvo**

**Input:** Task $\mathcal{T}$; initial design $\mathbf{x}_0$ and valid coverage $C_0$; physics prior $P$; full-wave evaluator $E$; round budget $B$; candidate counts $n_d, n_g$.

**Output:** Best valid design $\mathbf{x}^{\text{best}}$; round records $\mathcal{L}$; task-goal outcome.

$\mathcal{H} \leftarrow \emptyset$; $\mathcal{L} \leftarrow \emptyset$
$\mathbf{x}^{\text{ref}} \leftarrow \mathbf{x}_0$; $\mathbf{x}^{\text{best}} \leftarrow \mathbf{x}_0$
**for** $r = 1, \ldots, B$ **do**
    $\pi_r \leftarrow \text{LLMPLAN}(\mathcal{T}, \mathbf{x}^{\text{ref}}, \mathcal{H};\ \mathbf{x}_0, C_0)$
    $\mathcal{C}_r \leftarrow \text{BUILDPOOL}(\mathcal{T}, \mathbf{x}^{\text{ref}}, \pi_r, \mathcal{L};\ n_d, n_g)$
    **if** $\mathcal{C}_r = \emptyset$ **then**
        $\text{RECORDSKIPPED}(\mathcal{L}, r, \pi_r)$; **continue**
    **end if**
    $\mathbf{x}_r \leftarrow \text{TOPRANK}(P, \mathcal{C}_r, \mathcal{T})$
    $y_r \leftarrow E(\mathcal{T}, \mathbf{x}_r)$
    $\text{RECORD}(\mathcal{L}, r, \pi_r, \mathcal{C}_r, \mathbf{x}_r, y_r)$
    **if** $\text{VALID}(y_r, \mathcal{T})$ **then**
        $\mathcal{H} \leftarrow \mathcal{H} \cup \{(r, \mathbf{x}_r, y_r)\}$
        $\mathbf{x}^{\text{ref}} \leftarrow \text{BESTPROPOSAL}(\mathcal{H})$
        $\mathbf{x}^{\text{best}} \leftarrow \text{BETTER}(\mathbf{x}^{\text{best}}, \mathbf{x}_r)$
    **end if**
**end for**
**return** $\mathbf{x}^{\text{best}}, \mathcal{L}, \text{GOALOUTCOME}(\mathcal{H}, \mathcal{T})$

**Notes.** BuildPool combines plan-directed increments and global samples, applies the prescribed task constraints, mapping and deduplication rules, and returns $\emptyset$ for an invalid plan or incomplete pool. Skipped rounds still count toward $B$.

$\mathcal{H}$ contains valid new proposals only; $\mathcal{L}$ records all rounds. BestProposal selects the highest Exact coverage, with the earliest proposal retained under ties. Better compares the saved Exact coverage of its two inputs and retains the earlier design under ties. GoalOutcome tests new valid proposals only.

This study uses $B = 5$, $n_d = 36$, $n_g = 12$. Candidate generation and ranking follow the rules above; proposal-budget and retry rules are given in Section S2.3.

### S2.3. Proposal budget and retries

Each task receives five proposal opportunities. A round with a noncompliant plan or an incomplete candidate pool still consumes one opportunity. It does not proceed to Exact or receive a replacement proposal, and its coverage is recorded as missing.

The selected candidate is evaluated separately at each specified incidence angle. Each candidate–angle pair is allowed at most 10 attempts, including the initial attempt; retries stop once the evaluation succeeds. Geometry, materials and evaluation conditions remain unchanged during retries. Proposals, candidate–angle pairs and actual evaluation attempts are counted separately. Interrupted executions are resumed or checked before their outcomes are determined; they are not counted as design failures.

All five rounds are completed even if the target is reached earlier, and valid results below the target remain in the evaluation record. Task success requires at least one new proposal to pass physical validity and task-constraint checks and meet the target. The first-success round is the earliest round satisfying these conditions. If none succeeds, success is recorded as not observed within the five-round budget. The final best coverage may still come from Initial.

**S2.4. Independent two-variable illustration**

The independent two-variable MAHSS example (Supplementary Fig. 1) fixes the Debye parameters at $\varepsilon_\infty$ = 18.171, $\varepsilon_s$ = 206.118 and τ = 0.092 ns. Only coating thickness b (0.005–0.302 mm) and core height h (3–19.200 mm) are adjusted. The initial design has b = 0.264 mm and h = 18.302 mm, with coverage of 0.481. The target is coverage of 0.950 at an absorptance threshold of 0.900 over 2–18 GHz under 0° TE incidence. This example is independent of the 24-task held-out comparison.

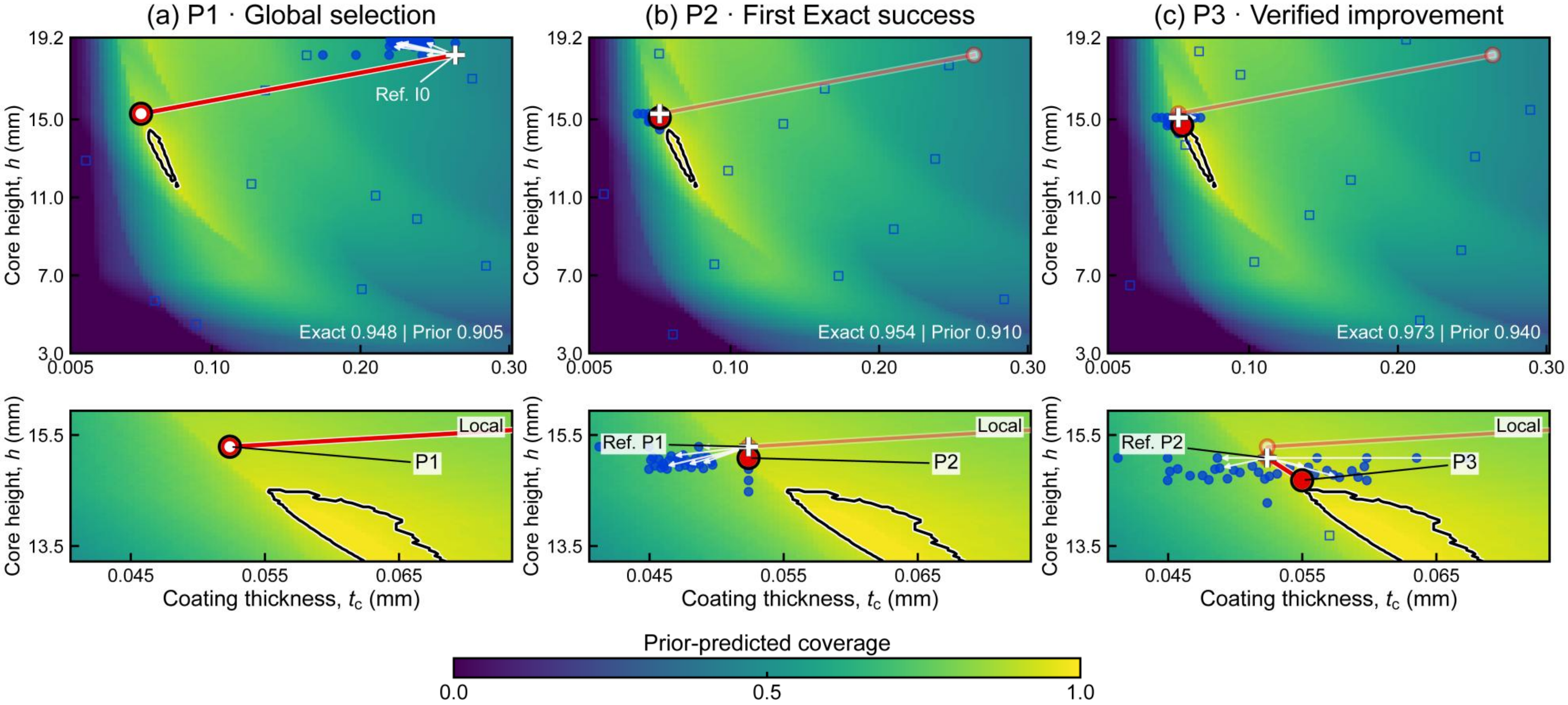

**Supplementary Fig. 1 | Candidate generation in the two-variable MAHSS example. (a–c) Proposals P1–P3 in the full domain (top) and local window (bottom). Squares: 12 global candidates; blue circles: 36 directed candidates; plus: generation reference; ring: top-ranked Prior candidate. White arrows show directed displacements. Filled and open red circles mark Exact outcomes meeting and missing the target. Background colours and the 0.950 contour show post-run Prior predictions.**

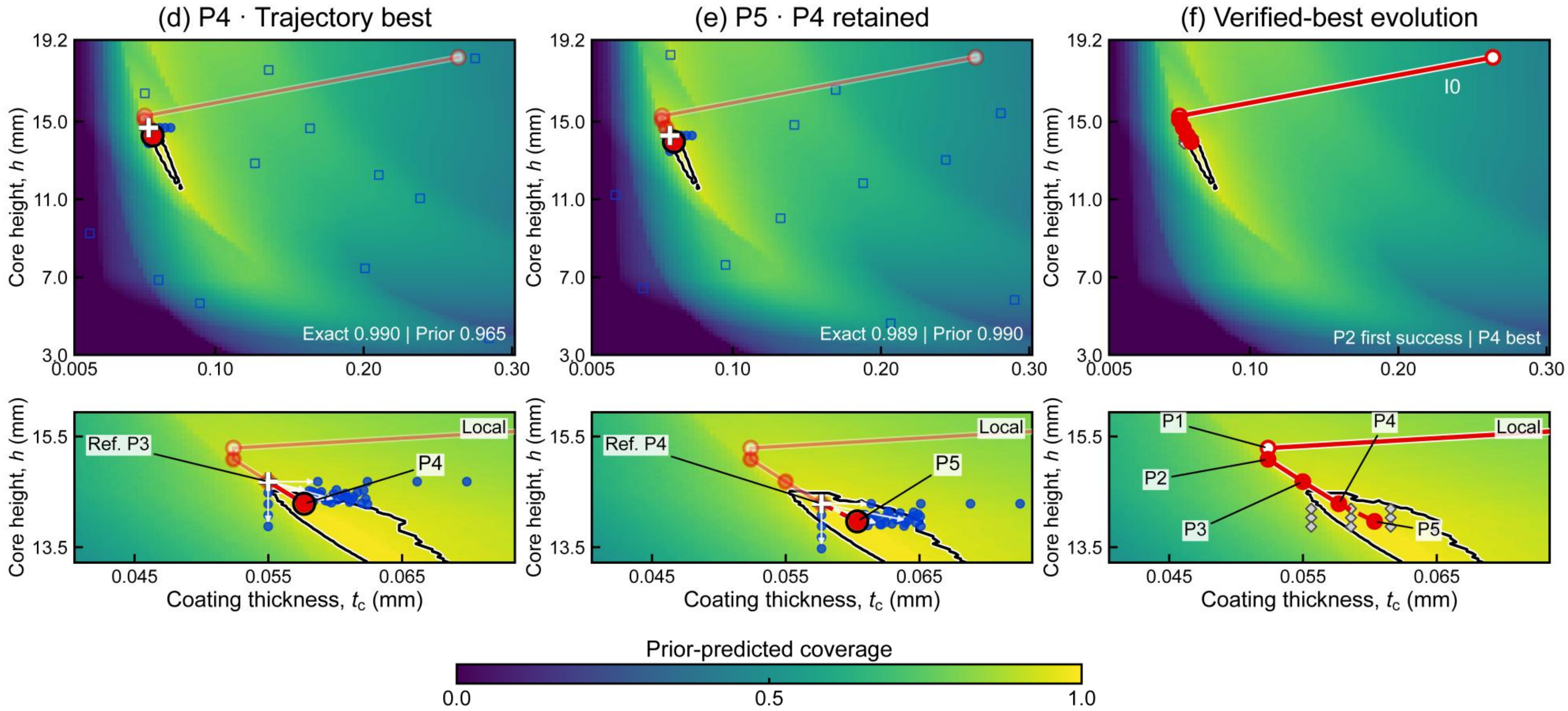

**Supplementary Fig. 1 (continued) | (d–e) Proposals P4–P5, with symbols as in (a–c). (f) Verified-best trajectory: P2 first meets the target, P4 is the best of five proposals, and the dashed branch marks P5. Grey diamonds denote pre-validation points W and N1–N8, withheld from the agent. Here, $t_c$ denotes the one-sided coating thickness b.**

### S2.5. System-level evaluation configurations

Supplementary Table 5 defines the configurations used in system-level evaluation. B0–B3 were compared on eight held-out tasks, comprising seven full-wave tasks and one offline error-diagnosis task. A0–A4 were evaluated on five independent ablation tasks. All configurations were scored by the same evaluator. These tests use different protocols from the five-proposal AbsorbBench comparison; removing runtime evaluator feedback does not remove final evaluation.

**Supplementary Table 5 |** Configurations for system-level evaluation.

| ID | Configuration | Definition / change |
|---|---|---|
| **System configuration comparison · 8 held-out tasks** | | |
| **B0** | Single-shot inference | One inference pass for task solving |
| **B1** | Execution retry | B0 with retry after execution failure |
| **B2** | C1 without feedback | Evolved skill C1; no runtime evaluator feedback |
| **B3** | C1 with feedback | Evolved skill C1 with runtime evaluator feedback |
| **Single-component ablation · 5 independent tasks** | | |
| **A0** | Full configuration | Reference configuration with evolved skill C1 |
| **A1** | No runtime feedback | A0 without runtime evaluator feedback |
| **A2** | No range checks | A0 without parameter-range checks |
| **A3** | No physics diagnosis | A0 without physics diagnosis |
| **A4** | Initial skill | A0 with initial skill S0 replacing C1 |

**Notes.** A1–A4 each modify one setting relative to A0; all other settings remain unchanged. All ablation variants are scored by the same evaluator. Removing runtime feedback does not remove final evaluation. S0 and C1 denote the initial and evolved skills, respectively.

## S3. AbsorbBench-36 task specifications

### S3.1. Tasks and data splits

AbsorbBench-36 comprises 18 MAHSS and 18 TPMS tasks, divided into 8 development, 4 validation and 24 held-out tasks. The formal comparison uses the 24 held-out tasks, with 12 from each structural family. Supplementary Table 6 lists target bands, incidence conditions, reflection loss (RL) thresholds and required coverage; Supplementary Table 7 gives design domains and initial designs.

Task objectives, design ranges, shared initial designs and data splits were fixed before the formal comparison and were not adjusted to the methods' results.

**Supplementary Table 6 | AbsorbBench-36 objectives: MAHSS tasks.**

| Task ID | Split | Band (GHz) | Incidence | RL threshold (dB) | Target coverage |
|---|---|---|---|---|---|
| AB20-HC-D01 | Development | 2–18 | 0°; TE and TM | -10 | 0.950 |
| AB20-HC-D02 | Development | 6–18 | 0°, 30°, 60°; TE and TM | -10 | 0.890 |
| AB20-HC-D03 | Development | 2–18 | 0°, 30°, 60°; TE and TM | -10 | 0.855 |
| AB20-HC-D04 | Development | 2–18 | 30°; TE | -10 | 0.950 |
| AB20-HC-T01 | Held-out | 2–18 | 60°; TE | -10 | 0.510 |
| AB20-HC-T02 | Held-out | 6–18 | 60°; TM | -11.500 | 0.950 |
| AB20-HC-T03 | Held-out | 2–18 | 0°, 30°; TE and TM | -10 | 0.940 |
| AB20-HC-T04 | Held-out | 8–18 | 30°, 60°; TE and TM | -10 | 0.855 |
| AB20-HC-V01 | Validation | 6–18 | 0°, 30°; TM | -10 | 0.950 |
| AB20-HC-V02 | Validation | 8–18 | 60°; TE and TM | -10 | 0.715 |
| AB36-HC-T05 | Held-out | 2–18 | 0°; TE | -10 | 0.950 |
| AB36-HC-T06 | Held-out | 6–18 | 30°; TE | -10 | 0.945 |
| AB36-HC-T07 | Held-out | 8–18 | 60°; TE | -10 | 0.585 |
| AB36-HC-T08 | Held-out | 2–18 | 0°; TM | -10 | 0.615 |
| AB36-HC-T09 | Held-out | 6–18 | 30°; TM | -10.500 | 0.900 |
| AB36-HC-T10 | Held-out | 2–18 | 30°, 60°; TM | -11.500 | 0.950 |
| AB36-HC-T11 | Held-out | 2–18 | 0°, 60°; TE and TM | -10 | 0.765 |
| AB36-HC-T12 | Held-out | 6–18 | 0°, 30°, 60°; TE and TM | -10.500 | 0.860 |

**Supplementary Table 6 (continued) | AbsorbBench-36 objectives: TPMS tasks.**

| Task ID | Split | Band (GHz) | Incidence | RL threshold (dB) | Target coverage |
|---|---|---|---|---|---|
| AB20-TPMS-D01 | Development | 5–16 | 45°; TE and TM | -5 | 0.925 |
| AB20-TPMS-D02 | Development | 5–16 | 50°; TE and TM | -4.500 | 0.925 |
| AB20-TPMS-D03 | Development | 5–18 | 45°, 50°; TE and TM | -5 | 0.915 |
| AB20-TPMS-D04 | Development | 5–16 | 20°; TE | -6.500 | 0.835 |
| AB20-TPMS-T01 | Held-out | 5–18 | 70°; TE | -5 | 0.395 |
| AB20-TPMS-T02 | Held-out | 5–16 | 10°, 50°; TM | -5.500 | 0.845 |
| AB20-TPMS-T03 | Held-out | 5–18 | 0°, 30°, 60°; TE and TM | -5.500 | 0.860 |
| AB20-TPMS-T04 | Held-out | 5–16 | 10°, 40°, 70°; TE and TM | -5.500 | 0.750 |
| AB20-TPMS-V01 | Validation | 5–18 | 40°; TM | -5 | 0.950 |
| AB20-TPMS-V02 | Validation | 5–16 | 50°; TE and TM | -5.500 | 0.650 |
| AB36-TPMS-T05 | Held-out | 8–18 | 60°; TE | -5 | 0.315 |
| AB36-TPMS-T06 | Held-out | 2–12 | 20°, 60°; TM | -5 | 0.500 |
| AB36-TPMS-T07 | Held-out | 4–14 | 40°; TM | -5 | 0.905 |
| AB36-TPMS-T08 | Held-out | 4–14 | 10°, 50°; TM | -5 | 0.885 |
| AB36-TPMS-T09 | Held-out | 4–14 | 40°; TE and TM | -5.500 | 0.840 |
| AB36-TPMS-T10 | Held-out | 5–18 | 0°, 30°, 50°; TE and TM | -5.500 | 0.820 |
| AB36-TPMS-T11 | Held-out | 5–18 | 0°, 20°, 40°; TE and TM | -6 | 0.900 |
| AB36-TPMS-T12 | Held-out | 5–16 | 10°, 30°, 50°; TE and TM | -5.500 | 0.805 |

The specified polarization applies at every listed angle. "TE and TM" requires both polarizations. Evaluation conditions remain fixed throughout the comparison.

For PEC-backed structures, an RL threshold L gives an absorptance threshold of $1 - 10^{L/10}$: 0.684 for −5 dB and 0.900 for −10 dB.

## S3.2. Design domains and initial designs

Only the variables listed in Supplementary Table 7 are adjusted for each task; all other conditions remain fixed.

**Supplementary Table 7 | Design domains and initial designs: MAHSS tasks.**

| Task ID | Variable ranges | Fixed settings / materials | Initial / coverage |
|---|---|---|---|
| AB20-HC-D01 | b: 0.005–0.400 mm; h: 3–30 mm<br>$\varepsilon_\infty$: 5–25; $\varepsilon_s$: 180–240; τ: 0.050–0.100 ns | Cell angle 60°; s = 2.550 mm; w = 0.055 mm | b = 0.016 mm; h = 15.107 mm<br>$\varepsilon_\infty$ = 11.041; $\varepsilon_s$ = 214.965<br>τ = 0.072 ns; $C_0$ = 0.219 |
| AB20-HC-D02 | b: 0.005–0.500 mm; h: 3–30 mm<br>$\varepsilon_\infty$: 5–24; $\varepsilon_s$: 180–240; τ: 0.050–0.100 ns | Cell angle 60°; s = 2.550 mm; w = 0.051 mm | b = 0.016 mm; h = 15.736 mm<br>$\varepsilon_\infty$ = 16.378; $\varepsilon_s$ = 197.248<br>τ = 0.090 ns; $C_0$ = 0.214 |
| AB20-HC-D03 | b: 0.010–0.440 mm; h: 5–28 mm<br>$\varepsilon_\infty$: 7.500–21.500; $\varepsilon_s$: 190–234; τ: 0.052–0.096 ns | Cell angle 60°; s = 2.850 mm; w = 0.048 mm | b = 0.387 mm; h = 26.485 mm<br>$\varepsilon_\infty$ = 20.709; $\varepsilon_s$ = 231.692<br>τ = 0.058 ns; $C_0$ = 0.150 |
| AB20-HC-D04 | b: 0.005–0.500 mm; h: 3–30 mm<br>$\varepsilon_\infty$: 5–25; $\varepsilon_s$: 180–240; τ: 0.050–0.100 ns | Cell angle 60°; s = 2.550 mm; w = 0.055 mm | b = 0.389 mm; h = 10.233 mm<br>$\varepsilon_\infty$ = 13.528; $\varepsilon_s$ = 229.487<br>τ = 0.070 ns; $C_0$ = 0.176 |
| AB20-HC-T01 | b: 0.005–0.500 mm; h: 3–30 mm<br>$\varepsilon_\infty$: 5–17; $\varepsilon_s$: 180–216; τ: 0.050–0.080 ns | Cell angle 60°; s = 2.550 mm; w = 0.055 mm | b = 0.443 mm; h = 3.490 mm<br>$\varepsilon_\infty$ = 13.348; $\varepsilon_s$ = 212.805<br>τ = 0.052 ns; $C_0$ = 0 |
| AB20-HC-T02 | b: 0.335–0.500 mm; h: 3–30 mm<br>$\varepsilon_\infty$: 13–25; $\varepsilon_s$: 204–240; τ: 0.070–0.100 ns | Cell angle 60°; s = 2.550 mm; w = 0.051 mm | b = 0.450 mm; h = 25.368 mm<br>$\varepsilon_\infty$ = 19.953; $\varepsilon_s$ = 221.084<br>τ = 0.078 ns; $C_0$ = 0 |
| AB20-HC-T03 | b: 0.104–0.401 mm; h: 8.400–24.600 mm<br>$\varepsilon_\infty$: 5–25; $\varepsilon_s$: 180–240; τ: 0.050–0.100 ns | Cell angle 60°; s = 2.850 mm; w = 0.048 mm | b = 0.267 mm; h = 20.592 mm<br>$\varepsilon_\infty$ = 20.547; $\varepsilon_s$ = 192.548<br>τ = 0.055 ns; $C_0$ = 0.374 |
| AB20-HC-T04 | b: 0.005–0.500 mm; h: 3–30 mm<br>$\varepsilon_\infty$: 9–21; $\varepsilon_s$: 192–228; τ: 0.060–0.090 ns | Cell angle 60°; s = 2.550 mm; w = 0.055 mm | b = 0.384 mm; h = 27.146 mm<br>$\varepsilon_\infty$ = 15.894; $\varepsilon_s$ = 199.664<br>τ = 0.065 ns; $C_0$ = 0.364 |
| AB20-HC-V01 | b: 0.203–0.500 mm; h: 3–19.200 mm<br>$\varepsilon_\infty$: 9–21; $\varepsilon_s$: 192–228; τ: 0.060–0.090 ns | Cell angle 60°; s = 2.850 mm; w = 0.048 mm | b = 0.444 mm; h = 8.650 mm<br>$\varepsilon_\infty$ = 11.870; $\varepsilon_s$ = 194.532<br>τ = 0.069 ns; $C_0$ = 0.415 |

**Supplementary Table 7 (continued) | Design domains and initial designs: MAHSS tasks.**

| Task ID | Variable ranges | Fixed settings / materials | Initial / coverage |
|---|---|---|---|
| AB20-HC-V02 | b: 0.203–0.500 mm; h: 13.800–30 mm<br>$\varepsilon_\infty$: 5–25; $\varepsilon_s$: 180–240; τ: 0.050–0.100 ns | Cell angle 60°; s = 2.850 mm; w = 0.048 mm | b = 0.467 mm; h = 23.650 mm<br>$\varepsilon_\infty$ = 7.874; $\varepsilon_s$ = 197.083<br>τ = 0.080 ns; $C_0$ = 0.374 |
| AB36-HC-T05 | b: 0.005–0.302 mm; h: 3–19.200 mm<br>$\varepsilon_\infty$: 5–25; $\varepsilon_s$: 180–240; τ: 0.050–0.100 ns | Cell angle 60°; s = 2.550 mm; w = 0.055 mm | b = 0.264 mm; h = 18.302 mm<br>$\varepsilon_\infty$ = 22.229; $\varepsilon_s$ = 210.460<br>τ = 0.051 ns; $C_0$ = 0.285 |
| AB36-HC-T06 | b: 0.104–0.401 mm; h: 8.400–24.600 mm<br>$\varepsilon_\infty$: 9–21; $\varepsilon_s$: 192–228; τ: 0.060–0.090 ns | Cell angle 60°; s = 2.550 mm; w = 0.051 mm | b = 0.120 mm; h = 19.294 mm<br>$\varepsilon_\infty$ = 12.921; $\varepsilon_s$ = 216.288<br>τ = 0.079 ns; $C_0$ = 0.853 |
| AB36-HC-T07 | b: 0.005–0.302 mm; h: 13.800–30 mm<br>$\varepsilon_\infty$: 9–21; $\varepsilon_s$: 192–228; τ: 0.060–0.090 ns | Cell angle 60°; s = 2.850 mm; w = 0.048 mm | b = 0.043 mm; h = 28.405 mm<br>$\varepsilon_\infty$ = 16.143; $\varepsilon_s$ = 214.160<br>τ = 0.063 ns; $C_0$ = 0.506 |
| AB36-HC-T08 | b: 0.203–0.500 mm; h: 13.800–30 mm<br>$\varepsilon_\infty$: 5–25; $\varepsilon_s$: 180–240; τ: 0.050–0.100 ns | Cell angle 60°; s = 2.550 mm; w = 0.051 mm | b = 0.351 mm; h = 24.156 mm<br>$\varepsilon_\infty$ = 10.500; $\varepsilon_s$ = 233.171<br>τ = 0.100 ns; $C_0$ = 0.402 |
| AB36-HC-T09 | b: 0.203–0.500 mm; h: 3–19.200 mm<br>$\varepsilon_\infty$: 9–21; $\varepsilon_s$: 192–228; τ: 0.060–0.090 ns | Cell angle 60°; s = 2.550 mm; w = 0.051 mm | b = 0.345 mm; h = 19.119 mm<br>$\varepsilon_\infty$ = 9.086; $\varepsilon_s$ = 225.600<br>τ = 0.064 ns; $C_0$ = 0.371 |
| AB36-HC-T10 | b: 0.005–0.500 mm; h: 3–30 mm<br>$\varepsilon_\infty$: 5–25; $\varepsilon_s$: 180–240; τ: 0.050–0.100 ns | Cell angle 60°; s = 2.850 mm; w = 0.048 mm | b = 0.047 mm; h = 19.474 mm<br>$\varepsilon_\infty$ = 23.158; $\varepsilon_s$ = 190.087<br>τ = 0.076 ns; $C_0$ = 0.865 |
| AB36-HC-T11 | b: 0.104–0.401 mm; h: 8.400–24.600 mm<br>$\varepsilon_\infty$: 9–21; $\varepsilon_s$: 192–228; τ: 0.060–0.090 ns | Cell angle 60°; s = 2.550 mm; w = 0.055 mm | b = 0.400 mm; h = 12.437 mm<br>$\varepsilon_\infty$ = 18.685; $\varepsilon_s$ = 209.557<br>τ = 0.072 ns; $C_0$ = 0.230 |
| AB36-HC-T12 | b: 0.005–0.500 mm; h: 3–30 mm<br>$\varepsilon_\infty$: 9–21; $\varepsilon_s$: 192–228; τ: 0.060–0.090 ns | Cell angle 60°; s = 2.850 mm; w = 0.048 mm | b = 0.441 mm; h = 14.488 mm<br>$\varepsilon_\infty$ = 13.518; $\varepsilon_s$ = 223.422<br>τ = 0.075 ns; $C_0$ = 0.260 |

**Supplementary Table 7 (continued) | Design domains and initial designs: TPMS tasks.**

| Task ID | Variable ranges | Fixed settings / materials | Initial / coverage |
|---|---|---|---|
| AB20-TPMS-D01 | t: 0.220–0.420; a: 10–20 mm | Topology: Primitive, Gyroid, Diamond<br>Material: M2, M3, M4, M5, M6, M7, M8 | Diamond; M6; t = 0.235; a = 10.568 mm<br>$C_0$ = 0.163 |
| AB20-TPMS-D02 | t: 0.220–0.420; a: 10–20 mm | Topology: Primitive, Gyroid, Diamond<br>Material: M2, M3, M4, M5, M6, M7, M8 | Gyroid; M4; t = 0.364; a = 12.116 mm<br>$C_0$ = 0.195 |
| AB20-TPMS-D03 | t: 0.220–0.420; a: 10–20 mm | Topology: Primitive, Gyroid, Diamond<br>Material: M2, M3, M4, M5, M6, M7, M8 | Diamond; M3; t = 0.224; a = 18.466 mm<br>$C_0$ = 0.228 |
| AB20-TPMS-D04 | VF: 0.300–0.700; a: 12–18 mm; VF step 0.010 | Topology: Primitive, Gyroid, Diamond<br>Material: M2, M3, M4, M5, M6, M7, M8 | Diamond; M3; VF = 0.340; a = 16.992 mm<br>$C_0$ = 0.139 |
| AB20-TPMS-T01 | VF: 0.100–0.550; a: 10–15 mm; VF step 0.010 | Topology: Primitive, Gyroid, Diamond<br>Material: M2, M3, M4, M5 | Diamond; M3; VF = 0.280; a = 10.876 mm<br>$C_0$ = 0.050 |
| AB20-TPMS-T02 | VF: 0.100–0.550; a: 12–18 mm; VF step 0.010 | Topology: Primitive, Gyroid, Diamond<br>Material: M2, M3, M4, M5 | Primitive; M2; VF = 0.100; a = 13.250 mm<br>$C_0$ = 0 |
| AB20-TPMS-T03 | VF: 0.300–0.700; a: 15–20 mm; VF step 0.010 | Topology: Primitive, Gyroid, Diamond<br>Material: M3, M4, M5, M6, M7 | Primitive; M3; VF = 0.640; a = 18.785 mm<br>$C_0$ = 0.711 |
| AB20-TPMS-T04 | VF: 0.100–0.900; a: 10–15 mm; VF step 0.010 | Topology: Gyroid<br>Material: M2, M3, M4, M5, M6, M7, M8 | Gyroid; M4; VF = 0.190; a = 13.100 mm<br>$C_0$ = 0.271 |
| AB20-TPMS-V01 | VF: 0.100–0.550; a: 15–20 mm; VF step 0.010 | Topology: Primitive, Gyroid, Diamond<br>Material: M2, M3, M4, M5 | Gyroid; M2; VF = 0.520; a = 15.946 mm<br>$C_0$ = 0.305 |
| AB20-TPMS-V02 | VF: 0.450–0.900; a: 10–15 mm; VF step 0.010 | Topology: Primitive, Gyroid, Diamond<br>Material: M5, M6, M7, M8 | Gyroid; M7; VF = 0.740; a = 10.258 mm<br>$C_0$ = 0.521 |
| AB36-TPMS-T05 | VF: 0.550–0.900; a: 10–14 mm; VF step 0.010 | Topology: Gyroid<br>Material: M6, M7, M8 | Gyroid; M6; VF = 0.760; a = 10.541 mm<br>$C_0$ = 0.127 |
| AB36-TPMS-T06 | VF: 0.100–0.450; a: 15–20 mm; VF step 0.010 | Topology: Diamond<br>Material: M2, M3, M4 | Diamond; M2; VF = 0.330; a = 15.683 mm<br>$C_0$ = 0 |

**Supplementary Table 7 (continued) | Design domains and initial designs: TPMS tasks.**

| Task ID | Variable ranges | Fixed settings / materials | Initial / coverage |
|---|---|---|---|
| AB36-TPMS-T07 | VF: 0.100–0.500; a: 15–20 mm; VF step 0.010 | Topology: Diamond<br>Material: M3, M4, M5, M6 | Diamond; M4; VF = 0.290; a = 15.413 mm<br>$C_0$ = 0.452 |
| AB36-TPMS-T08 | VF: 0.100–0.450; a: 15–20 mm; VF step 0.010 | Topology: Gyroid<br>Material: M3, M4, M5, M6 | Gyroid; M5; VF = 0.310; a = 15.337 mm<br>$C_0$ = 0.643 |
| AB36-TPMS-T09 | VF: 0.100–0.450; a: 15–20 mm; VF step 0.010 | Topology: Gyroid<br>Material: M3, M4, M5, M6 | Gyroid; M5; VF = 0.360; a = 15.903 mm<br>$C_0$ = 0.649 |
| AB36-TPMS-T10 | VF: 0.450–0.900; a: 10–15 mm; VF step 0.010 | Topology: Primitive, Gyroid, Diamond<br>Material: M3, M4, M5, M6, M7 | Gyroid; M7; VF = 0.640; a = 12.345 mm<br>$C_0$ = 0.815 |
| AB36-TPMS-T11 | VF: 0.100–0.550; a: 12–18 mm; VF step 0.010 | Topology: Gyroid<br>Material: M5, M6, M7, M8 | Gyroid; M7; VF = 0.270; a = 12.439 mm<br>$C_0$ = 0.686 |
| AB36-TPMS-T12 | VF: 0.450–0.900; a: 10–20 mm; VF step 0.010 | Topology: Primitive, Gyroid, Diamond<br>Material: M4, M5, M6, M7 | Gyroid; M5; VF = 0.870; a = 11.339 mm<br>$C_0$ = 0.585 |

MAHSS: b, one-sided coating thickness; h, core height; s, cell-wall side length; w, wall thickness; $\varepsilon_\infty$ and $\varepsilon_s$, relative Debye permittivities. Lengths are in mm and τ in ns. Layers and materials: Supplementary Tables 1–2.

TPMS: a, period; VF, target solid volume fraction; t, dimensionless level half-width. All tasks use a $64^3$ grid, one period per axis, xyz orientation and PEC backing. VF is mapped to t as described in S1.2. Materials M2–M8: Supplementary Table 2.

Values are rounded to three decimal places for display; calculations use the original parameters.

## S4. Skill evolution and transfer settings

### S4.1. Training, validation and transfer tasks

The independent skill evaluation includes 10 training, 6 validation and 8 held-out tasks (Supplementary Table 8). Training proceeded in batches of 4, 3 and 3 tasks. The cumulative experience from 4, 7 and 10 tasks produced textual guidance C1, C2 and C3, respectively. Eligible candidates were compared with the current version on the same validation tasks. C1 was selected, and its text was frozen before held-out testing. In this independent evaluation, the low-fidelity tool was used only for parameter-range checks.

Held-out tasks form three groups: same-type, cross-family and out-of-distribution (OOD). OOD conditions involve changes in TPMS geometry, material choices and MAHSS coating-thickness ranges. This evaluation is separate from the 24 optimization tasks in AbsorbBench-36.

**Supplementary Table 8 | Skill training tasks and success criteria.**

| Task ID | Stage | Task class | Conditions / design | Success criterion |
|---|---|---|---|---|
| TR-B-T01 | Train 1 | Build / TPMS | Gyroid; M1; level 0.260 | G; complete evidence |
| TR-D-H01 | Train 1 | Debug / MAHSS | Paired material-reference fault | Diagnose + minimal repair |
| TR-I-T01 | Train 1 | Improve / TPMS | Gyroid M1; level 0.260 → [0.270, 0.320] | G; Mean ≥ 0.053; Peak ≥ 0.125 |
| TR-I-H01 | Train 1 | Improve / MAHSS | H-B1; coating 0.035 → [0.040, 0.055] mm | G; Mean ≥ 0.940; Peak ≥ 0.980 |
| TR-B-H01 | Train 2 | Build / MAHSS | H-B1 | G; complete evidence |
| TR-D-T01 | Train 2 | Debug / TPMS | Paired frequency-unit fault | Diagnose + minimal repair |
| TR-I-T02 | Train 2 | Improve / TPMS | Gyroid; level 0.310; M1 → {M1, M2, M3, M4} | G; Mean ≥ 0.100; Peak ≥ 0.200 |
| TR-D-T02 | Train 3 | Debug / TPMS | Paired Floquet-mode fault | Diagnose + minimal repair |
| TR-D-H02 | Train 3 | Debug / MAHSS | Paired PEC-backing fault | Diagnose + minimal repair |
| TR-I-H02 | Train 3 | Improve / MAHSS | LFH-02; (tw, l, tc, h) ∈ [0.020, 0.060] × [2.500, 3.500] × [0.050, 0.085] × [12, 16] mm | G; Mean ≥ 0.940; Peak ≥ 0.980 |

The geometry tuple (tw, l, tc, h) gives wall thickness, cell-edge length, coating thickness and core height in mm. H-B1 = (0.035, 2.750, 0.035, 15); H-B2 = (0.035, 2.750, 0.050, 15).
MAHSS reference designs, rounded to three decimals: LFH-01 = (0.025, 2.707, 0.074, 14.048); LFH-02 = (0.058, 2.326, 0.027, 11.366); LFH-04 = (0.055, 3.174, 0.077, 12.031).

LFH cells use a 60° cell angle, 1.000 mm skins and 0.200 mm adhesive layers on both sides.

For skill tasks, the coating Debye parameters ($\varepsilon_\infty$, $\varepsilon_s$, $\tau$) are (5.019, 235.913, 0.090 ns). Wall, skin and adhesive permittivities are 3.800, 3.300 and 3.400, respectively; all have tan δ = 0.015.

**Supplementary Table 8 (continued) | Skill validation and held-out transfer tasks.**

| Task ID | Stage | Task class / group | Conditions / design | Success criterion |
|---|---|---|---|---|
| VA-B-T01 | Validation | Build / TPMS | Gyroid; M2; level 0.310 | G; complete evidence |
| VA-B-H01 | Validation | Build / MAHSS | LFH-04 | G; complete evidence |
| VA-D-M01 | Validation | Debug / Mixed | Paired polarization-column mapping fault | Diagnose + minimal repair |
| VA-I-T01 | Validation | Improve / TPMS | Gyroid M2; level 0.310 → [0.340, 0.380] | G; Mean ≥ 0.140; Peak ≥ 0.240 |
| VA-I-H01 | Validation | Improve / MAHSS | H-B1; (tw, l, tc, h) ∈ [0.020, 0.030] × [2.600, 2.800] × [0.065, 0.080] × [13.500, 14.500] mm | G; Mean ≥ 0.940; Peak ≥ 0.980 |
| VA-X-B01 | Validation | Transfer / TPMS | Gyroid → Diamond; level 0.330; {M1, M2} | G; Mean ≥ 0.100; Peak ≥ 0.200 |
| TE-B-T01 | Held-out | Same-type | Diamond; M2; level 0.330 | G; complete evidence |
| TE-B-H01 | Held-out | Same-type | LFH-01 | G; complete evidence |
| TE-I-H01 | Held-out | Same-type | H-B2; coating 0.050 → [0.065, 0.080] mm | G; Mean ≥ 0.940; Peak ≥ 0.980 |
| TE-D-M01 | Held-out | Cross-family | Mixed normalized export; energy-closure fault | Diagnose + minimal repair |
| TE-X-M01 | Held-out | Cross-family | Gyroid M1 → H-B2; target coating [0.065, 0.080] mm | G; Mean ≥ 0.940; Peak ≥ 0.980 |
| TE-I-T01 | Held-out | OOD condition | Diamond M2; level 0.270 → [0.300, 0.340] | G; Mean ≥ 0.100; Peak ≥ 0.200 |
| TE-X-T01 | Held-out | OOD condition | Diamond; level 0.330; {M1, M2, M3} | G; Mean ≥ 0.100; Peak ≥ 0.200 |
| TE-X-H01 | Held-out | OOD condition | H-B2; coating [0.085, 0.100] mm | G; Mean ≥ 0.950; Peak ≥ 0.995 |

All tasks use 2–18 GHz, normal incidence, TE/TM and PEC backing. TPMS uses a 10 mm period and $64^3$ geometry grid; level denotes dimensionless sheet half-width. Build and Debug assess construction and diagnosis, respectively.

G requires Executable, PhysicsValid and Converged status. Debug requires correct diagnosis and minimal repair with protected fields fixed. Over the task band, Mean = min(mean A_TE, mean A_TM) and Peak = min(max A_TE, max A_TM).

Success is evaluated using the thresholds listed in the table. EAB and the additional relative-gain condition for TR-I-T01 are excluded from the success score.

### S4.2. Skill content and selection

Skill evolution summarizes design actions, evaluation outcomes and errors from training tasks into textual guidance, while keeping model weights, materials and solver settings unchanged. Validation tasks select the version, and held-out tasks do not contribute to updates. Building on S0's single-candidate and pre-solve checking rules, C1 adds material-reference checks, Floquet-mode prechecks and guidance for single-variable coating adjustments (Supplementary Table 9).

**Supplementary Table** 9 **| Changes from the initial skill S0 to the selected skill C1.**

| Aspect | Initial skill S0 | Selected skill C1 |
|---|---|---|
| Material references | Identify the violated condition and make a minimal repair. | Check paired material IDs and restore incorrect references. |
| Floquet-mode checks | Keep ports and modes fixed; stop when a check fails. | Check task-visible Floquet modes before solving; stop if missing or altered. |
| Bounded parameter adjustment | Propose one design within the permitted variables. | For coating-only tasks, change one thickness and hold other parameters fixed. |

Candidate skills are compared with the current version on the same 6 validation tasks. Physical validity, convergence and error diagnosis must not decline, and protected-field violations and information leakage are prohibited. Candidates are then compared in order of target attainment, task success, performance metrics and computational cost, with shorter skill text as the final preference. Replacement requires a strictly better candidate; otherwise, the current version is retained. C1 was selected after the first validation round and was not replaced by later candidates.

### S4.3. Transfer evaluation and cost accounting

S0 and C1 were run independently on the same eight held-out tasks, with one session and at most one design candidate per task. Debug tasks were completed offline. Existing full-wave results may be reused when geometry, materials, excitation and solver settings match, and are re-evaluated under the current task conditions. Proposal generation remains independent.

Full-wave-equivalent cost is calculated per task. The indicator $I_i$ identifies an Improve/Transfer task i, and $P_{iv}$ indicates that the proposal from version v passes pre-solve checks. The indicator $J_{iv}$ denotes use of a new full-wave evaluation or an identical-input result already generated during this test. Each indicator equals 1 when its condition holds and 0 otherwise. The cost is

$$K_v = \sum_{i=1}^{8} I_i P_{iv} J_{iv} \tag{S11}$$

Each eligible design task counts as one full-wave-equivalent cost unit. If both versions reuse the same result generated during this test, each version incurs one unit. Results available before testing are excluded, as are Build, offline Debug and tasks that fail pre-solve checks. The cost $K_v$ does not depend on target attainment. It measures comparable task-level cost, not the actual number of solver calls or elapsed time.

S0 and C1 each completed 8 task attempts and incurred 4 full-wave-equivalent cost units under these rules. Transfer outcomes are classified by comparing both versions on each task: failure-to-success, success-to-failure, success under both skills, or failure under both skills.